\pdfoutput=1

\documentclass[11pt]{article}

\usepackage{latex/acl}
\usepackage{makecell}
\usepackage{multirow}
\usepackage{float}
\usepackage{amssymb}
\usepackage{multirow}
\usepackage{color}
\usepackage{xspace}
\usepackage{subcaption}
\usepackage{times}
\usepackage{latexsym}
\usepackage{amsmath}
\usepackage{tabularx}
\usepackage{booktabs}
\usepackage{array}
\usepackage{tikz}
\usepackage{colortbl}
\usepackage{xcolor}
\usepackage{authblk}
\DeclareMathOperator*{\argmaxA}{arg\,max}
\usepackage[T1]{fontenc}

\usepackage[utf8]{inputenc}

\usepackage{microtype}

\usepackage{inconsolata}

\usepackage{graphicx}

\usepackage{times}  
\usepackage{algorithm}
\usepackage{algorithmic}
\usepackage{ulem}

\title{Large Language Models Are Better Logical Fallacy Reasoners with Counterargument, Explanation, and Goal-Aware Prompt Formulation}

\author{
    Jiwon Jeong$^{1}$ \quad 
    Hyeju Jang$^{2}$\thanks{Corresponding authors.} \quad 
    Hogun Park$^{1}$\protect\footnotemark[1] \\
    $^{1}$Sungkyunkwan University, Republic of Korea \\
    $^{2}$Indiana University Indianapolis, USA \\
    \texttt{jwjw9603@skku.edu} \quad
    \texttt{hyejuj@iu.edu} \quad
    \texttt{hogunpark@skku.edu}
}

\begin{document}
\maketitle
\newcommand{\modelLLaMA}{LLaMA\xspace}
\newcommand{\modelGPT}{GPT\xspace}

\begin{abstract}\label{abstract}
The advancement of Large Language Models (LLMs) has greatly improved our ability to process complex language. However, accurately detecting logical fallacies remains a significant challenge. This study presents a novel and effective prompt formulation approach for logical fallacy detection, applicable in both supervised (fine-tuned) and unsupervised (zero-shot) settings. Our method enriches input text incorporating implicit contextual information---\textit{counterarguments}, \textit{explanations}, and \textit{goals}---which we query for validity within the context of the argument. We then rank these queries based on confidence scores to inform classification. We evaluate our approach across multiple datasets from 5 domains, covering 29 distinct fallacy types, using models from the \modelGPT and \modelLLaMA series. The results show substantial improvements over state-of-the-art models, with F1 score increases of up to 0.60 in zero-shot settings and up to 0.45 in fine-tuned models. Extensive analyses further illustrate why and how our method excels.  
\end{abstract}

\section{Introduction}\label{introduction}

Logical fallacies are flawed arguments resulting from faulty reasoning ~\cite{Clark1971-CLAROC-4}. For example, consider the statement: ``Annie must like Starbucks because all girls like Starbucks.'' This exemplifies the logical fallacy known as \textit{faulty generalization}, as it assumes that all girls universally share a preference for Starbucks---an overly broad generalization. 
Recognizing logical fallacies is essential for both humans and AI. It enables us to avoid being misled by faulty arguments, improves communication by minimizing errors, prevents misleading or deceptive misinformation, and strengthens convincing arguments \cite{woods2004death, tindale2007fallacies}. 

\begin{figure}[!t]
    \centering
    \begin{minipage}{\linewidth}
        \centering
        \includegraphics[width=\linewidth]{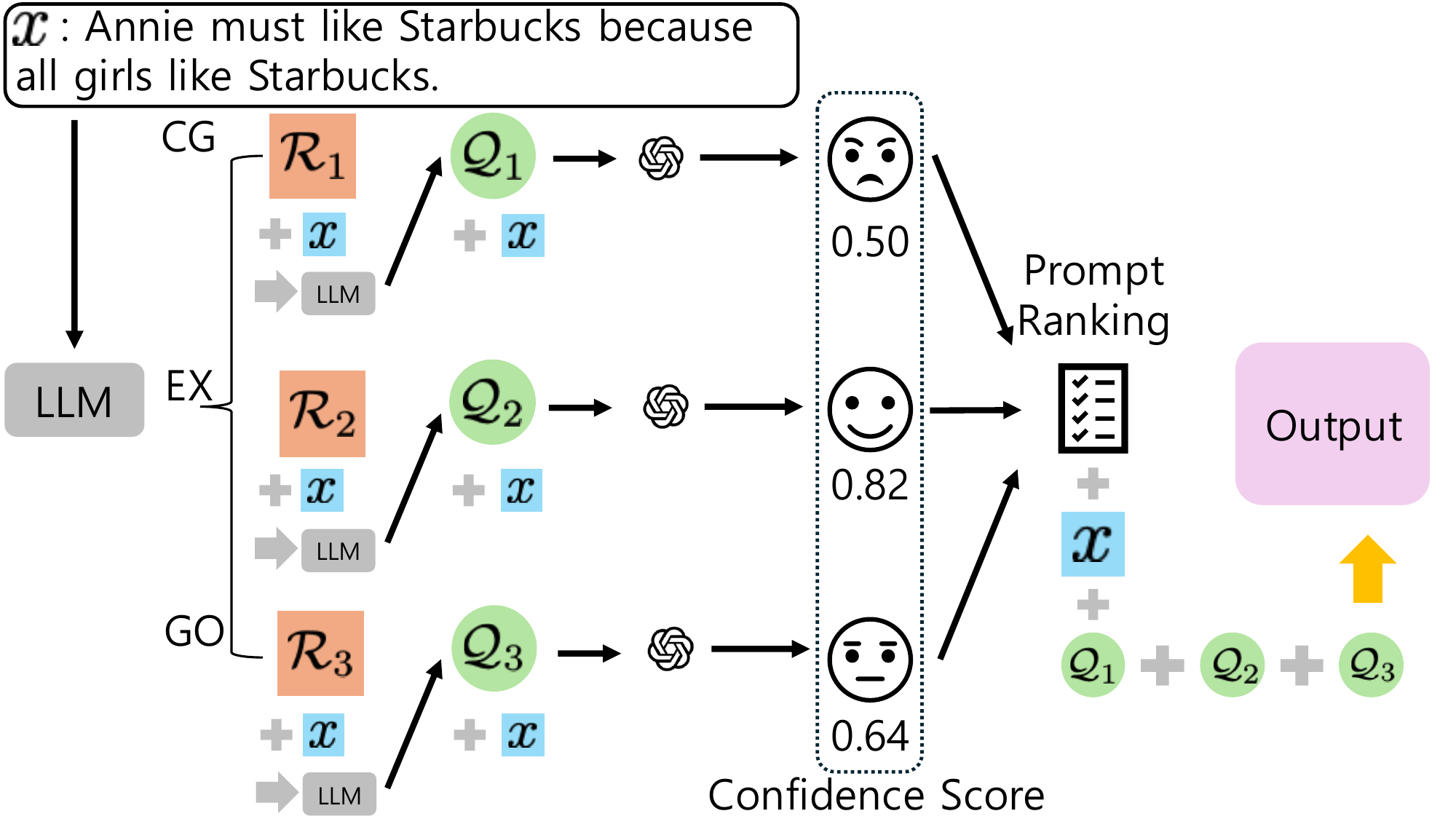}

    \end{minipage}
    \vspace{+2mm}
    \begin{minipage}{\linewidth}
        \centering
        \includegraphics[width=\linewidth]{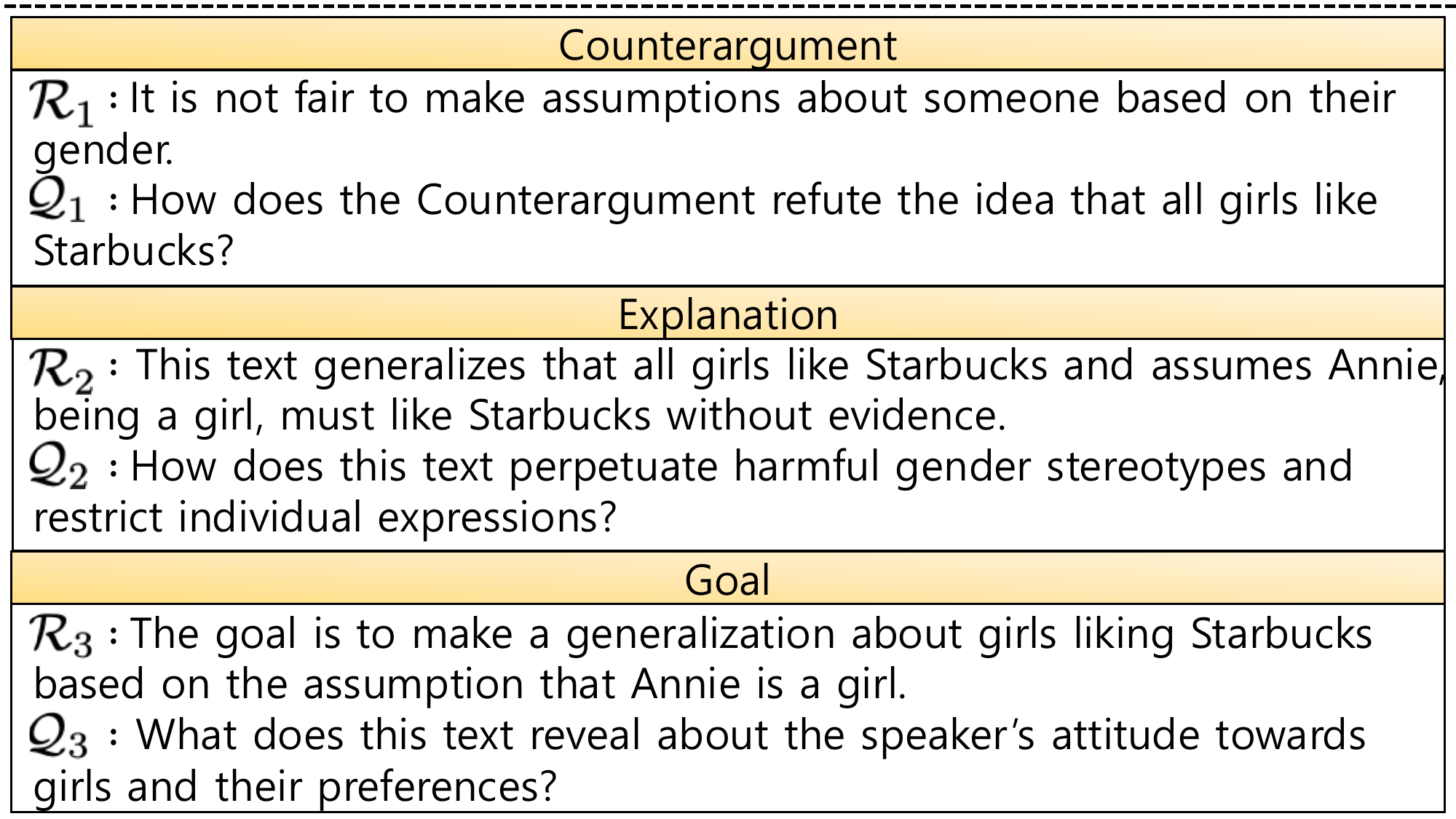} 
        
    \end{minipage}
    \caption{Prompt formulation: \(x\) represents the input text to classify. \( \mathcal{R}_i\) denotes the contextual augmentation generated from the input text using specific instructions for Counterargument (CG), Explanation (EX), and Goal (GO). \( \mathcal{Q}_i\) denotes the reformulated queries created from each augmentation to analyze the input text.}
    \label{fig:model}
    \vspace{-2mm}
\end{figure}

Despite this importance, logical fallacy detection remains in its early stage in natural language processing. Recent large language models (LLMs) have demonstrated challenges not only in general reasoning~\cite{naveed2023comprehensive,chen2024premise} but also in detecting logical fallacies. For example, \citet{jin2022logical} showed that both zero-shot and fine-tuning results of BERT-based models are suboptimal. More recently, \citet{hong2024closer} highlighted that most LLMs continue to struggle in this area. To the best of our knowledge, there has been no attempt to address this issue through prompting engineering or other unsupervised methods.

To address this gap, we investigate whether incorporating implicit information from multiple perspectives on arguments into prompts can improve LLM performance in logical fallacy detection. We propose a novel two-step prompt formulation process that leverages implicit contextual information (See Figure~\ref{fig:model}), applicable in both supervised (fine-tuned) and unsupervised (zero-shot) settings. Inspired by \citet{sourati2023case}, we leverage three key types of implicit information relevant to arguments: \textit{Counterargument},  \textit{Explanation}, and \textit{Goal}. We hypothesize that this implicit information provides additional context for identifying logical fallacies.

To operationalize this idea, we augment input text in the prompt with this implicit contextual information. Specifically, we begin by generating statements summarizing the relevant background (counterargument, explanation, and goal), followed by queries that assess the validity of this information within the argument's context. These queries are then ranked using a prompt ranking method based on confidence scores. Then, this ranking information is integrated into the input prompts. Notably, all steps of this process are automatically generated, requiring no manual prompt adjustments.

We evaluate our approach on five publicly available datasets, covering 29 distinct fallacy types, using models from the \modelGPT and \modelLLaMA series.  Our experiments show that this approach substantially outperforms baselines and state-of-the-art models in both unsupervised and supervised settings. Specifically, compared to state-of-the-art models, we achieve up to an improvement of 0.60 in Macro-F1 score in zero-shot settings and 0.45 in fine-tuned models. Although \citet{sourati2023case} pioneered the use of this type of information, how we extract and use the information differs substantially, resulting in superior performance. We also perform extensive analyses to explore why and how our method excels, including examining the impact of our prompt ranking method, performing calibration analysis, and conducting error analysis.
Our code is available at \url{https://github.com/jw9603/Logical_Fallacy}.

Our contributions can be summarized in three ways. (1) We introduce a structured prompt design for logical fallacy detection that effectively integrates three types of implicit contextual information. This goes beyond previous methods by using a confidence-based prompt ranking system, which incorporates diverse information to enhance the model’s analysis. (2) We perform extensive evaluation across five datasets, demonstrating the effectiveness of our approach across various domains. (3) We conduct extensive in-depth analyses to validate our methodology, examining the effectiveness of each component in design.

\section{Related Work}\label{related_work}

\subsection{Logical Fallcy Detection}

Previous research explored computational methods for logical fallacy detection in various contexts, including dialogues \cite{habernal2017argotario}, argument sufficiency \cite{stab2017recognizing,wachsmuth2017computational}, Reddit discussions \cite{sahai2021breaking},  misinformation \cite{musi2022developing}, and educational materials \cite{jin2022logical}. These studies focused on single datasets, limiting their ability to demonstrate generalizability across diverse domains with varying argument structures, which is crucial for robust fallacy detection in real-world scenarios.

Most recent approaches aimed to address this issue by tackling multiple datasets. For instance, \citet{alhindi2022multitask} enhanced fallacy detection performance across multiple datasets by employing a T5 model \cite{raffel2020exploring} with multitask instruction-based prompting, outperforming single-dataset models. \citet{sourati2023case} proposed a  Case-Based Reasoning method, enriching cases with implicit information, including \textit{counterarguments}, \textit{goals}, \textit{explanations}, and \textit{structure}, to retrieve similar cases. While both approaches demonstrated effectiveness, they rely on supervised methods which require large datasets. 

Furthermore, \citet{hong2024closer} evaluated various LLMs using prompting techniques. They experimented with providing definitions of logical fallacies within sentences and without, but found little improvement in performance. This study revealed that LLMs exhibit limited performance in fallacy detection, emphasizing the need for specialized reasoning methods tailored to the task.

Similar to prior research, our study aims to detect logical fallacies across multiple datasets, but it is differentiated by its focus on leveraging LLMs more effectively in both supervised and unsupervised settings. We build on the idea of using implicit contextual information, \textit{counterarguments}, \textit{goals}, and \textit{explanations}, introduced by \citet{sourati2023case}. While our results corroborate their findings on the value of such information for the task, our work diverges significantly from theirs by focusing on prompt engineering rather than case-based retrieval. Although they demonstrated the utility of the implicit information for logical fallacy detection, their work does not explore how LLMs can generate or utilize this information. In contrast, our research systematically integrates these contextual elements into LLMs' reasoning process, thereby improving classification accuracy without requiring extensive labeled data. This shift is crucial, as it moves the focus from manual case-based reasoning to an automated, scalable solution leveraging LLMs' in-context learning capabilities.

\subsection{Prompting Method for Contextual Augmentation}

There has been similar work using prompts for contextual augmentation and query reformulation. \citet{shen-etal-2024-retrieval} refine queries using retrieved results to improve search performance. \citet{yang2023good} propose Question-Driven Visual Exploration (QVix), generating exploratory questions to enhance Large Vision-Language Models (LVLMs)’ ability to analyze visual content. While both approaches leverage LLMs to enhance query formulation or reasoning, they focus on retrieval optimization or visual understanding, respectively, rather than logical analysis. Unlike these prior approaches, our method constructs structured reasoning prompts specifically designed to facilitate logical fallacy classification, ensuring that LLMs engage in systematic argument evaluation.

Additionally, it is worth noting that our use of the term ``query'' differs from its role in information retrieval. In our context, ``query'' represents a question designed to guide logical fallacy classification, rather than a search input for document retrieval.

\section{Our Approach}\label{approach}

Our approach consists of four main steps to generate structured prompts. As shown in Figure \ref{fig:model}, we first use the LLM to create contextual augmentations, and based on these augmentations and the input text, we generate context-informed queries. Next, we classify logical fallacies using the generated queries, and finally, we rank the queries based on their confidence scores, incorporating this ranking information into the final classification.

We focus on three types of implicit information related to arguments for augmenting input text: \textit{Counterargument}, \textit{Explanation}, and \textit{Goal}. This contextual information provides diverse insights into logical fallacies, as demonstrated by \citet{sourati2023case}. Counterarguments present alternative viewpoints \cite{nussbaum2005effects}, explanations dissect the logic and reasoning \cite{hempel1948studies}, and goals evaluate whether the argument supports its conclusion \cite{tracy2013understanding}.

\paragraph{Step 1: Generate Contextual Augmentation}
 
First, we use LLMs to generate three types of implicit information, i.e., \textit{Counterargument}, \textit{Explanation}, and \textit{Goal}. Each type of information is denoted by an index \(i\), representing the specific instruction applied (\(\mathcal{I}^1\) for Counterargument, \(\mathcal{I}^2\) for Explanation, and \(\mathcal{I}^3\) for Goal). These instructions guide the LLM to generate contextual augmentations \(\mathcal{R}_i\). By applying specific instructions \(\mathcal{I}^i\) to \texttt{gpt-3.5-turbo-instruct}, we generate a  contextual augmentation:
\begin{equation}
    \label{equation1}
    \mathcal{R}_i = \text{LLM}(x, \mathcal{I}^i). 
\end{equation}

For instance, for the statement \(x =\) ``\textit{Annie must like Starbucks because all girls like Starbucks},'' the Goal perspective prompt is ``\textit{Express the goal of the text},'' leading to \(\mathcal{R}_i =\) \textit{``The goal of this text is to make a generalization about girls liking Starbucks based on the assumption that Annie is a girl.''} This contextually augmented text is used for generating a query.

\begin{table}[t]
\centering

\resizebox{\columnwidth}{!}{
\begin{tabular}{ccccc}

\hline

{\color[HTML]{000000} \textbf{Data}} & \textbf{N}    & \textbf{C}  & \textbf{Genre}       & \textbf{Domain}    \\ \hline \hline
PROPAGANDA                              & 12,267  & $16\dagger$ & News        & Politics   \\ \hline
ARGOTARIO                            & 1,338 & $6\dagger$  & Dialogue    & General   \\ \hline
LOGIC                                & 2,449 & 13 & Dialogue    & Education \\ \hline
COVID-19                             & 154  & $11\dagger$ & SocMed/News & Covid-19  \\ \hline
CLIMATE                              & 685  & $11\dagger$ & News        & Climate   \\ \hline
\end{tabular}}
\caption{Summary of five fallacy datasets. \textbf{N}: \# of samples, \textbf{C}: \# of classes. \textbf{$\dagger$}: No-Fallacy class included.}
\label{tab:dataset}
\end{table}
\paragraph{Step 2: Generate Reformulated Queries}

To generate context-informed queries, we design a tailored query generation method using the LLM instruction: ``\textit{Create one query for each text to analyze the text based on its goal rather than directly asking what a logical fallacy is.}'' This process yields \(\mathcal{Q}_{i} =\) ``\textit{What does this text reveal about the speaker's attitude towards girls and their preferences?}'' Such queries enable a nuanced understanding of the underlying assumptions and biases in arguments:
\vspace{-2mm}

\begin{equation}
    \label{equation2}
    \quad \mathcal{Q}_{i} = \text{LLM}(x, \mathcal{R}_i).
\end{equation}

\paragraph{Step 3: Calculate Confidence Scores for Queries}\label{step3}
In this step, we calculate the confidence scores for predicting the probability \(p_{\text{LLM}}\) of the fallacy's class label (\(l\)) with each contextual augmentation \(i\) for an input text \(x\):
\vspace{-2mm}
\begin{equation*}
    \text{conf}(\mathcal{Q}_i) = \max\limits_{l \in L} \sum_{\text{tokens}} \log p_{\text{LLM}}(l | x, \mathcal{Q}_i),
\end{equation*}

\noindent where $L$ denotes a set of logical fallacy labels. Here, \(\text{conf}(\mathcal{Q}_i)\) represents the highest confidence score derived from the log probabilities of the predicted tokens, indicating how confidently the model predicts the most probable class based on the specific query type.
The reformulated queries and their respective confidence scores will later be used in the following step to finalize the classification through ranking-based methods.

\paragraph{Step 4: Prompt Ranking-Based Classification}\label{step4}

To predict a final label considering all contexts, we use the confidence scores calculated in the previous step. Specifically, the confidence scores \(\text{conf}(\mathcal{Q}_i)\) for each query are sorted in descending order and incorporated into the final prompt as ranking information as follows:

\vspace{-2mm}
\begin{equation*}
    \label{equation4}
    \hat{y} = \argmaxA_{l \in L} p_{\text{LLM}}(l | x, Q_{all}, \text{Rank}(Q_{all})),
\end{equation*}

\noindent where $Q_{all} = \{ \mathcal{Q}_{i}, \forall i \}$. Here, \(\mathcal{Q}_i\) contains the actual content of each query (e.g., Counterargument, Explanation, Goal).
The function, $\text{Rank}(Q_{all})= \text{Sorted}(\{\text{conf}(\mathcal{Q}_i), \forall i\})$, ranks the queries based on their respective confidence scores. This ranking is then converted into text and included as part of the prompt. 
Both the content and the ranking information are utilized for enhancing classification performance. Further details on the prompting methods and examples can be found in Appendix~\ref{appendix:four_step}.

\section{Evaluation}\label{experiment}

\begin{table*}[!ht]
\centering
\resizebox{\textwidth}{!}{
\begin{tabular}{>{\centering\arraybackslash}m{3cm} >{\centering\arraybackslash}m{3cm} llllllllllllll}
\toprule
\textbf{LM} & \textbf{Supervised} & \textbf{Method} & \multicolumn{2}{c}{\textbf{PROPAGANDA}} & \multicolumn{2}{c}{\textbf{ARGOTARIO}} & \multicolumn{2}{c}{\textbf{LOGIC}} & \multicolumn{2}{c}{\textbf{COVID-19}} & \multicolumn{2}{c}{\textbf{CLIMATE}} \\ 
\cmidrule(lr){4-5} \cmidrule(lr){6-7} \cmidrule(lr){8-9} \cmidrule(lr){10-11} \cmidrule(lr){12-13}
 &  &  & \multicolumn{1}{c}{\textbf{ACC}} & \multicolumn{1}{c}{\textbf{F1}} & \multicolumn{1}{c}{\textbf{ACC}} & \multicolumn{1}{c}{\textbf{F1}} & \multicolumn{1}{c}{\textbf{ACC}} & \multicolumn{1}{c}{\textbf{F1}} & \multicolumn{1}{c}{\textbf{ACC}} & \multicolumn{1}{c}{\textbf{F1}} & \multicolumn{1}{c}{\textbf{ACC}} & \multicolumn{1}{c}{\textbf{F1}} \\ 
\midrule \midrule
\multirow{4}{*}{Fine-tuned LMs} & \multirow{4}{*}{\checkmark} & \cite{jin2022logical} & \multicolumn{1}{c}{---} & \multicolumn{1}{c}{---} & \multicolumn{1}{c}{---} & \multicolumn{1}{c}{---} & 0.48 & 0.59 & \multicolumn{1}{c}{---} & \multicolumn{1}{c}{---} & \multicolumn{1}{c}{---} & \multicolumn{1}{c}{---} \\
 &  & \cite{sourati2023case} & \multicolumn{1}{c}{\textbf{0.95}} & \multicolumn{1}{c}{\textbf{0.91}} & \multicolumn{1}{c}{0.71} & \multicolumn{1}{c}{0.71} & \multicolumn{1}{c}{0.76} & \multicolumn{1}{c}{0.71} & \multicolumn{1}{c}{0.36} & \multicolumn{1}{c}{0.18} & \multicolumn{1}{c}{0.48} & \multicolumn{1}{c}{0.29} \\
 &  & \cite{alhindi2022multitask} & \multicolumn{1}{c}{0.73} & \multicolumn{1}{c}{0.56} & \multicolumn{1}{c}{0.64} & \multicolumn{1}{c}{0.64} & \multicolumn{1}{c}{0.70} & \multicolumn{1}{c}{0.66} & \multicolumn{1}{c}{0.29} & \multicolumn{1}{c}{0.28} & \multicolumn{1}{c}{0.25} & \multicolumn{1}{c}{0.20} \\
 &  & \texttt{roberta-base} + EX (Ours) & \multicolumn{1}{c}{0.91} & \multicolumn{1}{c}{0.84} & \multicolumn{1}{c}{\textbf{0.81}} & \multicolumn{1}{c}{\textbf{0.80}} & \multicolumn{1}{c}{\textbf{0.80}} & \multicolumn{1}{c}{\textbf{0.79}} & \multicolumn{1}{c}{\textbf{0.36}} & \multicolumn{1}{c}{0.17} & \multicolumn{1}{c}{\textbf{0.84}} & \multicolumn{1}{c}{\textbf{0.74}} \\
 \midrule
\multirow{4}{*}{\texttt{gpt-3.5-turbo}} & \multirow{4}{*}{$\times$} & Zero-shot & \multicolumn{1}{c}{0.12} & \multicolumn{1}{c}{0.07} & \multicolumn{1}{c}{0.50} & \multicolumn{1}{c}{0.41} & \multicolumn{1}{c}{0.39} & \multicolumn{1}{c}{0.25} & \multicolumn{1}{c}{0.29} & \multicolumn{1}{c}{0.14} & \multicolumn{1}{c}{0.18} & \multicolumn{1}{c}{0.13} \\
 &  & ZCoT \cite{kojima2022large} & \multicolumn{1}{c}{0.14} & \multicolumn{1}{c}{0.08} & \multicolumn{1}{c}{0.48} & \multicolumn{1}{c}{0.39} & \multicolumn{1}{c}{0.39} & \multicolumn{1}{c}{0.23} & \multicolumn{1}{c}{0.29} & \multicolumn{1}{c}{0.11} & \multicolumn{1}{c}{0.20} & \multicolumn{1}{c}{0.13} \\
 &  & DEF \cite{hong2024closer} & \multicolumn{1}{c}{0.23} & \multicolumn{1}{c}{0.11} & \multicolumn{1}{c}{0.50} & \multicolumn{1}{c}{0.48} & \multicolumn{1}{c}{0.40} & \multicolumn{1}{c}{0.22} & \multicolumn{1}{c}{0.29} & \multicolumn{1}{c}{0.11} & \multicolumn{1}{c}{0.16} & \multicolumn{1}{c}{0.09} \\
 &  & Prompt Ranking (Ours) & \multicolumn{1}{c}{\textbf{0.35}} & \multicolumn{1}{c}{\textbf{0.17}} & \multicolumn{1}{c}{\textbf{\underline{0.84}}} & \multicolumn{1}{c}{\textbf{0.69}} & \multicolumn{1}{c}{\textbf{0.45}} & \multicolumn{1}{c}{\textbf{0.32}} & \multicolumn{1}{c}{\textbf{0.71}} & \multicolumn{1}{c}{\textbf{0.56}} & \multicolumn{1}{c}{\textbf{0.63}} & \multicolumn{1}{c}{\textbf{0.58}} \\
 \midrule
\multirow{4}{*}{\texttt{gpt-4}} & \multirow{4}{*}{$\times$} & Zero-shot & \multicolumn{1}{c}{0.32} & \multicolumn{1}{c}{0.19} & \multicolumn{1}{c}{0.60} & \multicolumn{1}{c}{0.50} & \multicolumn{1}{c}{0.40} & \multicolumn{1}{c}{0.28} & \multicolumn{1}{c}{0.36} & \multicolumn{1}{c}{0.25} & \multicolumn{1}{c}{0.18} & \multicolumn{1}{c}{0.11} \\
 &  & ZCoT \cite{kojima2022large} & \multicolumn{1}{c}{0.31} & \multicolumn{1}{c}{\textbf{\underline{0.20}}} & \multicolumn{1}{c}{0.56} & \multicolumn{1}{c}{0.47} & \multicolumn{1}{c}{0.41} & \multicolumn{1}{c}{0.28} & \multicolumn{1}{c}{0.43} & \multicolumn{1}{c}{0.33} & \multicolumn{1}{c}{0.17} & \multicolumn{1}{c}{0.12} \\
 &  & DEF \cite{hong2024closer} & \multicolumn{1}{c}{0.34} & \multicolumn{1}{c}{0.17} & \multicolumn{1}{c}{0.60} & \multicolumn{1}{c}{0.52} & \multicolumn{1}{c}{0.40} & \multicolumn{1}{c}{0.28} & \multicolumn{1}{c}{0.43} & \multicolumn{1}{c}{0.29} & \multicolumn{1}{c}{0.18} & \multicolumn{1}{c}{0.14} \\
 &  & Prompt Ranking (Ours) & \multicolumn{1}{c}{\textbf{\underline{0.40}}} & \multicolumn{1}{c}{\textbf{\underline{0.20}}} & \multicolumn{1}{c}{\textbf{\underline{0.84}}} & \multicolumn{1}{c}{\textbf{\underline{0.83}}} & \multicolumn{1}{c}{\textbf{\underline{0.49}}} & \multicolumn{1}{c}{\textbf{\underline{0.37}}} & \multicolumn{1}{c}{\textbf{\underline{0.86}}} & \multicolumn{1}{c}{\textbf{\underline{0.76}}} & \multicolumn{1}{c}{\textbf{\underline{0.78}}} & \multicolumn{1}{c}{\textbf{\underline{0.71}}} \\
 \midrule
\multirow{4}{*}{\texttt{Llama3-8B}} & \multirow{4}{*}{$\times$} & Zero-shot & \multicolumn{1}{c}{0.13} & \multicolumn{1}{c}{0.05} & \multicolumn{1}{c}{0.21} & \multicolumn{1}{c}{0.13} & \multicolumn{1}{c}{0.25} & \multicolumn{1}{c}{0.13} & \multicolumn{1}{c}{0.07} & \multicolumn{1}{c}{0.02} & \multicolumn{1}{c}{0.14} & \multicolumn{1}{c}{0.08} \\
 &  & ZCoT~\cite{kojima2022large} & \multicolumn{1}{c}{0.23} & \multicolumn{1}{c}{0.04} & \multicolumn{1}{c}{0.19} & \multicolumn{1}{c}{0.13} & \multicolumn{1}{c}{0.20} & \multicolumn{1}{c}{0.09} & \multicolumn{1}{c}{0.07} & \multicolumn{1}{c}{0.07} & \multicolumn{1}{c}{0.13} & \multicolumn{1}{c}{0.06} \\
 &  & DEF~\cite{hong2024closer} & \multicolumn{1}{c}{\textbf{0.28}} & \multicolumn{1}{c}{0.06} & \multicolumn{1}{c}{0.25} & \multicolumn{1}{c}{0.13} & \multicolumn{1}{c}{0.23} & \multicolumn{1}{c}{0.11} & \multicolumn{1}{c}{0.07} & \multicolumn{1}{c}{0.03} & \multicolumn{1}{c}{0.14} & \multicolumn{1}{c}{0.07} \\
 &  & Prompt Ranking (Ours) & \multicolumn{1}{c}{0.22} & \multicolumn{1}{c}{\textbf{0.09}} & \multicolumn{1}{c}{\textbf{0.30}} & \multicolumn{1}{c}{\textbf{0.29}} & \multicolumn{1}{c}{\textbf{0.38}} & \multicolumn{1}{c}{\textbf{0.36}} & \multicolumn{1}{c}{\textbf{0.36}} & \multicolumn{1}{c}{\textbf{0.25}} & \multicolumn{1}{c}{\textbf{0.33}} & \multicolumn{1}{c}{\textbf{0.40}} \\
 \bottomrule
\end{tabular}}
\caption{Multi-class fallacy classification performance. \textbf{Bold}: the highest score for each base model. \textbf{\underline{Bold}}: the highest score across all methods for each dataset in the unsupervised setting. Results for \cite{jin2022logical} and \cite{alhindi2022multitask} are taken from their original papers, with the same test splits used for comparison. \textbf{---}: The model requires additional annotations to run. \texttt{roberta-base} + EX (Ours) refers to our supervised approach using Explanation (EX) queries, which show the best performance among our proposed query types.
}
\label{tab:multiclass}
\end{table*}

To evaluate our method, we conduct experiments on multi-class classification to identify specific types of logical fallacies.

\subsection{Datasets}
We experiment with the five fallacy datasets: PROPAGANDA~\cite{da2019fine}, ARGOTARIO~\cite{habernal2017argotario}, LOGIC~\cite{jin2022logical}, COVID-19~\cite{musi2022developing}, and CLIMATE~\cite{alhindi2022multitask}.

These datasets cover 5 different domains and 29 unique fallacy types, as shown in Table~\ref{tab:dataset}. This breadth allows us to assess the generalizability of our method.

\subsection{Baselines}
We compare our method against several baselines in both unsupervised and supervised settings. In the unsupervised setting, we first use a baseline zero-shot approach that detects logical fallacies in a given text without using any additional information. We also compare our model against previous work, including Zero-shot CoT (ZCoT) \cite{kojima2022large}, which utilizes a Chain of Thought approach, and DEF \cite{hong2024closer}, which provides the model with fallacy definitions. Further details on ZCoT and DEF are in Appendix~\ref{appendix:zcot,def}. 

In the supervised setting, we compare our method against state-of-the-art models proposed by \citet{jin2022logical},  \citet{sourati2023case},  and \citet{alhindi2022multitask}. \citet{sourati2023case} employ contextual augmentations within a case-based reasoning framework, while \citet{alhindi2022multitask} apply multi-task learning for fallacy detection. 
 
\citet{jin2022logical} employs a structure-aware classifier that leverages additional coreference annotations to learn structural patterns in arguments, which is only applicable to the LOGIC dataset.
We report the results from the original papers for \cite{jin2022logical} and \cite{alhindi2022multitask}, and run the code provided by \citet{sourati2023case} to obtain their results.

\subsection{Model Implementation Details}

For the unsupervised setting, we use LLMs from the \modelGPT and
\modelLLaMA series with different sizes: \texttt{gpt-3.5-turbo}\footnote{We use \texttt{gpt-3.5-turbo-0125}.}, \texttt{gpt-4}\footnote{We use \texttt{gpt-4-0613}.}, \texttt{Llama2-7b-hf}, \texttt{Llama2-13b-hf}, and \texttt{Llama3-8B}. 

For the supervised setting, we implement our approach by fine-tuning a \texttt{roberta-base} model ~\cite{liu2019roberta} for fallacy classification. We concatenate the input text with each query as input to the model. The prompt ranking process is excluded in this setting, as it relies on confidence scores from LLMs like \texttt{gpt-3.5-turbo}. 
We train the \texttt{roberta-base} model using the AdamW optimizer \cite{loshchilov2019decoupled} on two NVIDIA RTX A6000 GPUs, which takes \textasciitilde 30 minutes. The batch size is chosen from \{4, 8, 16, 32, 64\} and the learning rate from \{1e-5, 2e-5\}. These hyperparameters are tuned using the development set.

\subsection{Results}
\begin{figure}[!t]
\centering
\includegraphics[width=\linewidth]{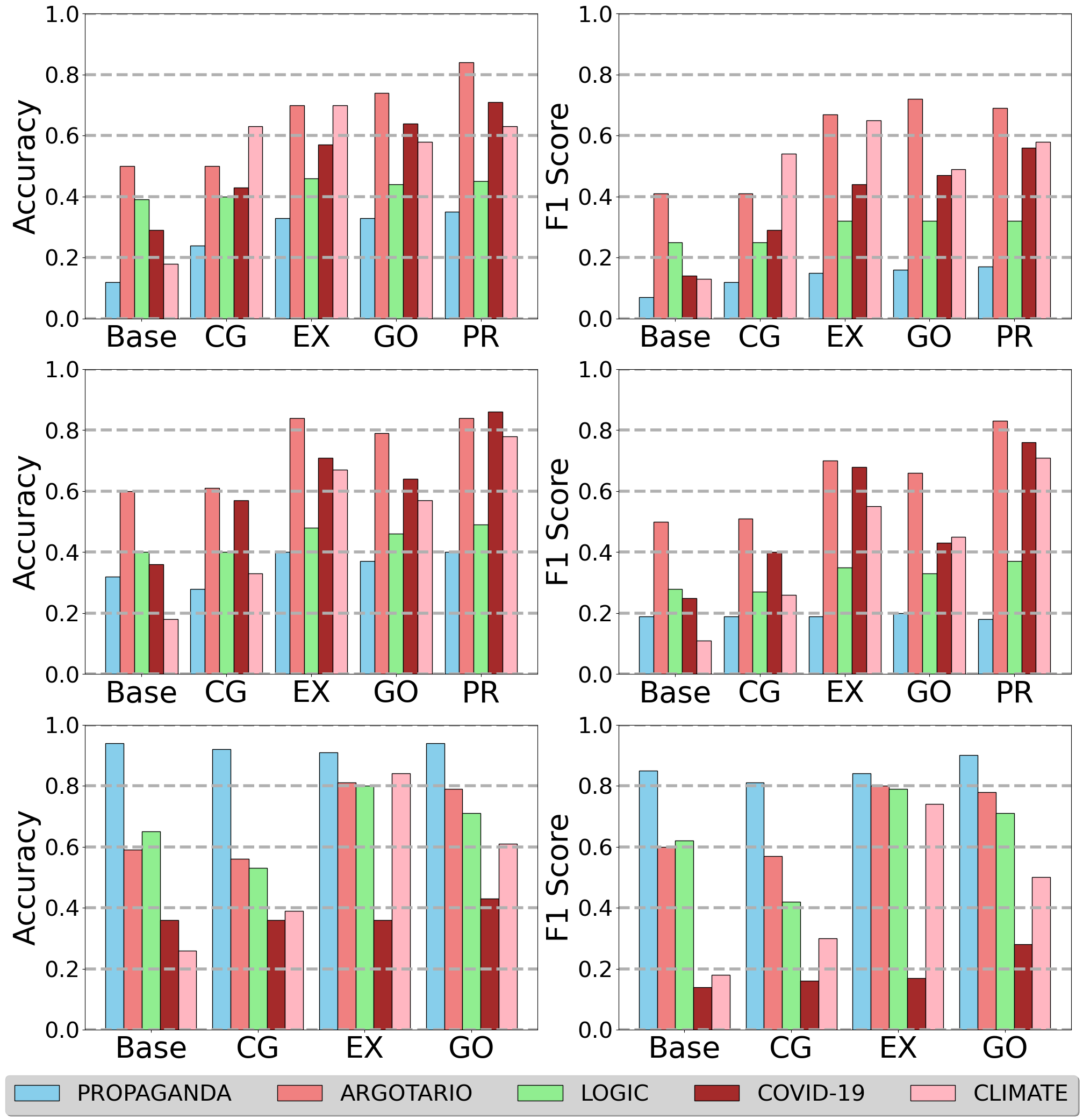}
\caption{Multi-class classification results based on query types for all datasets using \texttt{gpt-3.5-turbo}, \texttt{gpt-4}, and \texttt{roberta-base} from top to bottom. CG: Counterargument, EX: Explanation, GO: Goal, and PR: Prompt Ranking. Base: the method that uses only logical fallacy sentences without any queries.}
\label{fig:classification results based on reformulated queries}

\end{figure}

\paragraph{Overall Performance}
\label{result}
Table~\ref{tab:multiclass} shows the accuracy and Macro-F1 scores for multi-class fallacy classification across all datasets. The table includes the best-performing LLaMA model for the unsupervised setting in addition to two \modelGPT models, and the best-performing supervised model (\texttt{roberta-base} + EX) among our three queries. For comprehensive results, refer to  Appendix \ref{appendix:classification} and Figure \ref{fig:classification results based on reformulated queries}. 

Our prompt ranking method overall outperforms other approaches in unsupervised settings. For instance, \texttt{gpt-4} with prompt ranking achieves the highest scores on the PROPAGANDA dataset (accuracy: 0.40, Macro-F1: 0.20). In the supervised settings, \texttt{roberta-base} with Explanation queries shows the best performance on the CLIMATE dataset, achieving an accuracy of 0.84 and a Macro-F1 score of 0.74.

\paragraph{Performance per Query Type}\label{Classification Results Based on Reformulated Queries}

Figure~\ref{fig:classification results based on reformulated queries} shows the overall performance of each query type. A consistent trend is observed across datasets and models: Explanation (EX) queries yield the best overall performance, excelling across diverse datasets and achieving highest scores in both accuracy and F1 metrics. Goal (GO) queries also perform well, particularly in datasets like PROPAGANDA, where the content often has a clear objective or hidden intent, making the GO queries effective at identifying the underlying purpose behind persuasive arguments. However, Counterargument (CG) queries consistently underperform relative to EX and GO queries. While CG queries contribute to detection performance, their overall effectiveness is lower, particularly when compared to the stronger results achieved with EX and GO queries.

\begin{figure}[!t]
\centering
\includegraphics[width=\linewidth]{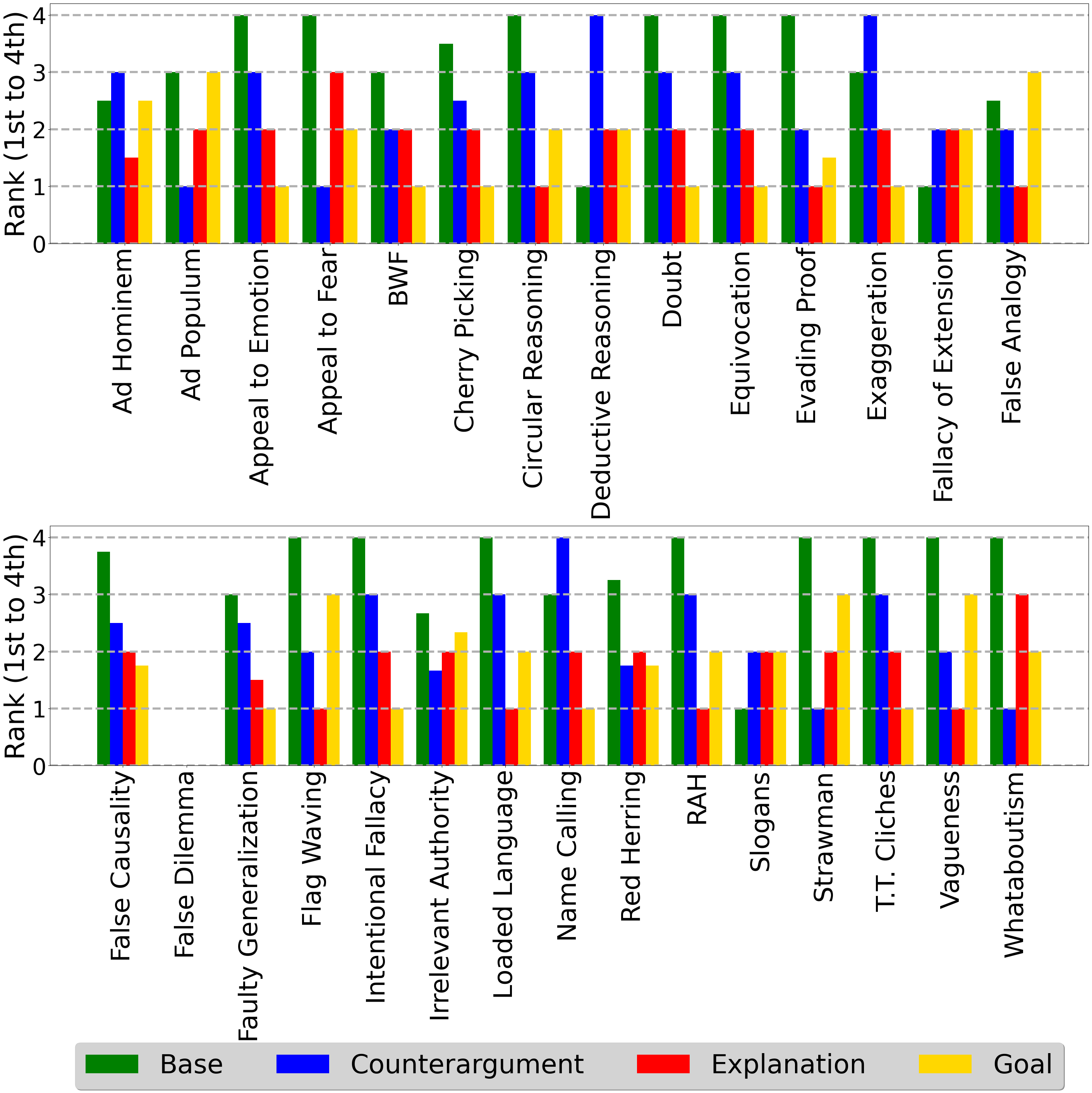}
\caption{Performance comparison of base (without queries) and all query types across various fallacy classes using the \texttt{gpt-3.5-turbo} model. The y-axis represents the average rank of each method across datasets. Lower ranks indicate better performance. BWF: Black and White Fallacy, RAH: Reductio Ad Hitlerum, T.T. Cliches: Thought Terminating Cliches.}

\label{fig:Comparative Performance of Reformulated Queries across Fallacy Class}
\end{figure}

Figure~\ref{fig:Comparative Performance of Reformulated Queries across Fallacy Class} shows the rankings of each query type across 29 fallacy classes.  EX queries consistently perform best, followed by GO queries, with CG queries ranking lowest. EX queries excel in fallacies requiring detailed reasoning, like False Analogy and Evading the Burden of Proof, while GO queries perform well in intent-driven fallacies, such as Appeal to Emotion and Intentional Fallacy. CG queries, though generally less effective, are suited for simpler logical fallacies like Ad Populum and Strawman.

Some exceptions stand out. In the Slogans fallacy, the Base method outperforms all query types, possibly because its simplicity is more effective for direct and emotionally charged statements. Similarly, in the Flag Waving fallacy, EX queries outperform GO, suggesting that identifying logical inconsistencies may sometimes be more effective than focusing on emotional intent. Overall, the queries consistently outperform the Base method, demonstrating their value in fallacy detection.

\section{Further Analyses}\label{analysis}

\subsection{Performance Based on Confidence Scores}\label{confidence}

\begin{figure}[!t]
\centering
\includegraphics[width=\linewidth]{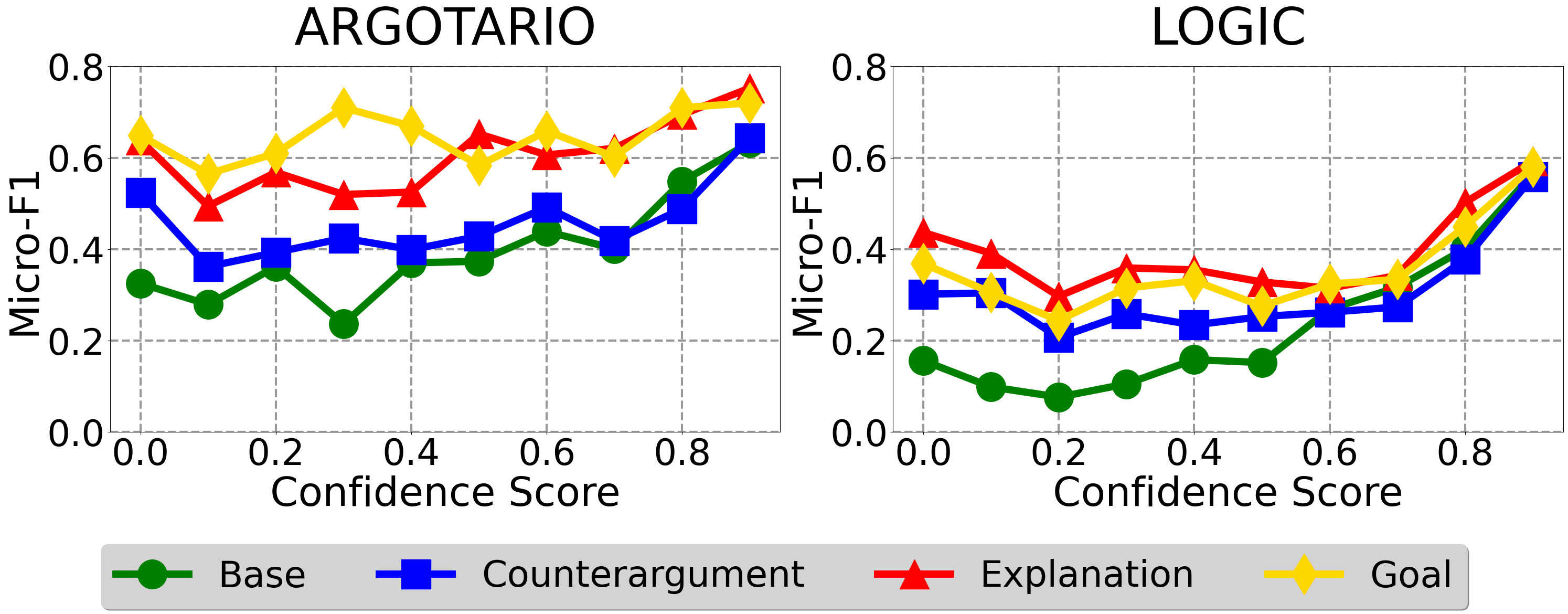}
\caption{Relationship between confidence scores and performance with/without queries for two datasets using the \texttt{gpt-3.5-turbo} model.}

\label{fig:confidence}
\end{figure}

We further investigate how the model's performance with each individual query type varies based on different confidence scores, which refer to the probability assigned by the LLM when generating tokens in the query, prior to the prompt ranking step. We measure Micro-F1 scores to assess overall performance across all classes, rather than focusing on class-level performance, using the \texttt{gpt-3.5-turbo} model. Figure~\ref{fig:confidence} illustrates the results for the ARGOTARIO and LOGIC datasets. The performance trends are consistent across other datasets.

The results indicate that incorporating queries consistently improves F1 scores across different confidence scores in all datasets. In particular, at lower confidence scores, the performance gap between our model and the baseline is more pronounced. This suggests that queries provide valuable contextual information, helping to clarify the underlying logic of the text.

\subsection{Calibration Analysis for Prompt Ranking}\label{calibration}

\begin{figure}[!t]
\centering

\includegraphics[width=\linewidth]{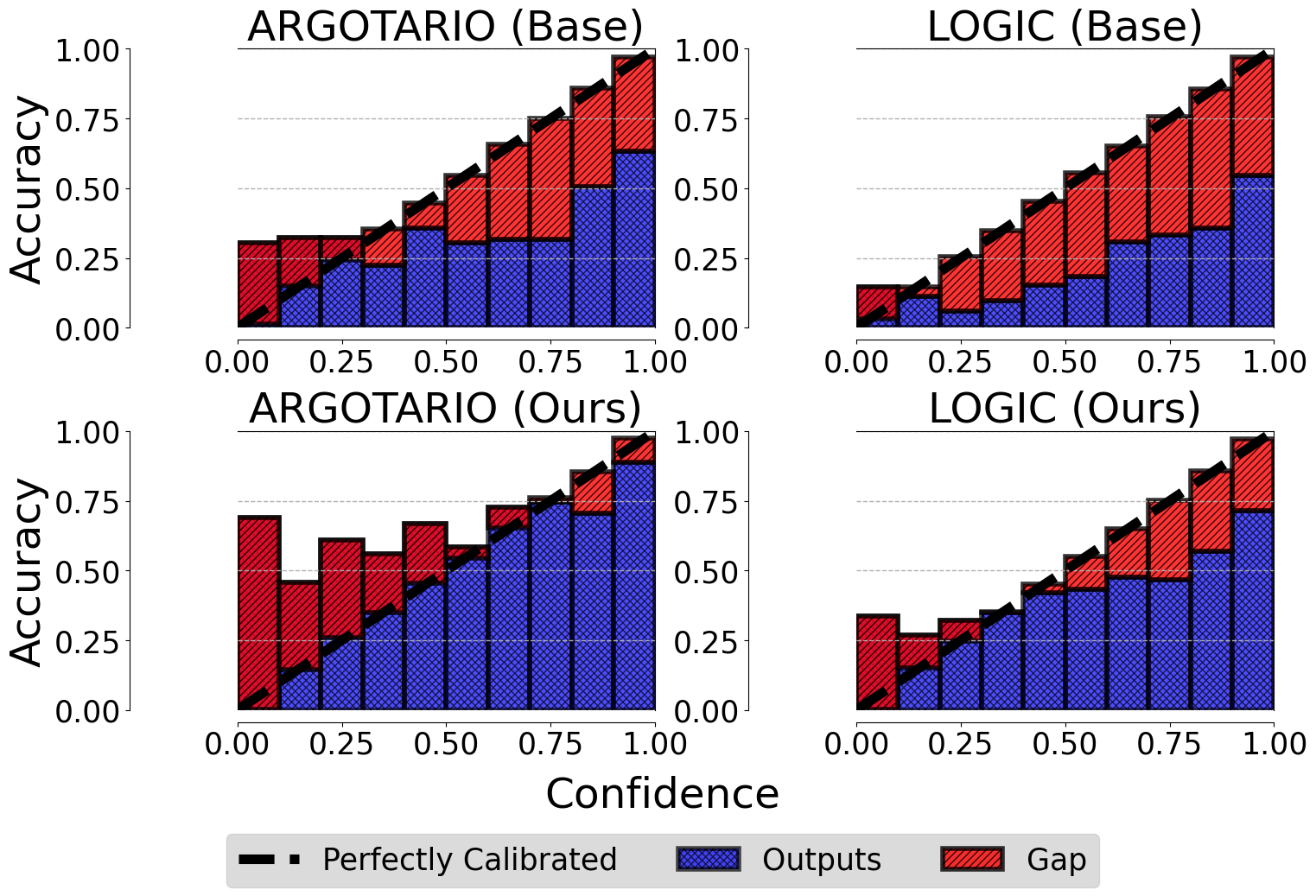}
\caption{Reliability diagrams comparing the calibration of the base method (without queries) and ours (prompt ranking) using the \texttt{gpt-3.5-turbo} model across two datasets.}

\label{fig:calibration}
\end{figure}

We aim to show that the Prompt Ranking (PR) method achieves better calibration compared to the Base method, which does not use queries, by evaluating both methods using the \texttt{gpt-3.5-turbo} model. Calibration is assessed using the Reliability Diagram~\cite{guo2017calibration}, which compares predicted confidence with observed accuracy, revealing deviations from perfect calibration, such as overconfidence or underconfidence. The calibration results for the ARGOTARIO and LOGIC datasets are presented in Figure~\ref{fig:calibration}.

Overall, while our PR demonstrates better calibration compared to the Base method, the LLMs still show some degree of overconfidence, consistent with findings from previous studies~\cite{yin2023large,ren2023investigating}. This trend is observed across other datasets such as PROPAGANDA, COVID-19, and CLIMATE, where PR consistently outperforms the Base method, further underscoring its effectiveness in leveraging multiple context-informed queries to provide a more nuanced interpretation of the input text and reduce confidence biases.
Since the final prediction is made by ranking the three confidence scores, our model demonstrates high accuracy even when confidence scores are low.

\begin{table*}[!ht]
\centering
\large
\resizebox{\textwidth}{!}{
\begin{tabular}{lllllllllllll}
\toprule
\textbf{LLM} & \textbf{Method} & \multicolumn{2}{c}{\textbf{PROPAGANDA}} & \multicolumn{2}{c}{\textbf{ARGOTARIO}} & \multicolumn{2}{c}{\textbf{LOGIC}} & \multicolumn{2}{c}{\textbf{COVID-19}} & \multicolumn{2}{c}{\textbf{CLIMATE}} \\ \cmidrule(lr){3-4} \cmidrule(lr){5-6} \cmidrule(lr){7-8} \cmidrule(lr){9-10} \cmidrule(lr){11-12}
 &  & \textbf{ACC} & \textbf{F1} & \textbf{ACC} & \textbf{F1} & \textbf{ACC} & \textbf{F1} & \textbf{ACC} & \textbf{F1} & \textbf{ACC} & \textbf{F1} \\ \midrule \midrule
\texttt{gpt-3.5-turbo} & None & 0.33 & 0.16 & 0.71 & 0.67 & 0.44 & 0.30 & 0.64 & 0.56 & 0.61 & 0.53 \\
 & Random & 0.33$_{\pm 0.01}$ & \textbf{0.17$_{\pm 0.01}$} & 0.76$_{\pm 0.02}$ & 0.63$_{\pm 0.06}$ & 0.41$_{\pm 0.01}$ & 0.29$_{\pm 0.01}$ & 0.58$_{\pm 0.05}$ & 0.43$_{\pm 0.06}$ & 0.31$_{\pm 0.03}$ & 0.25$_{\pm 0.03}$ \\
 & Prompt Ranking (Ours) & \textbf{0.35} & \textbf{0.17} & \textbf{\underline{0.84}} & \textbf{0.69} & \textbf{0.45} & \textbf{0.32} & \textbf{0.71} & \textbf{0.56} & \textbf{0.63} & \textbf{0.58} \\
  \midrule
\texttt{gpt-4} & None & 0.39 & 0.19 & 0.81 & 0.80 & 0.48 & 0.36 & 0.93 & 0.85 & 0.72 & 0.59 \\
 & Random &  0.40$_{\pm 0.01}$ & 0.19$_{\pm 0.02}$ & 0.83$_{\pm 0.01}$ & 0.77$_{\pm 0.06}$ & 0.48$_{\pm 0.01}$ & 0.36$_{\pm 0.01}$ & 0.79$_{\pm 0.03}$ & 0.71$_{\pm 0.06}$ & 0.31$_{\pm 0.03}$ & 0.25$_{\pm 0.01}$ \\
 & Prompt Ranking (Ours) & \textbf{\underline{0.40}} & \textbf{\underline{0.20}} & \textbf{\underline{0.84}} & \textbf{\underline{0.83}} & \textbf{\underline{0.49}} & \textbf{\underline{0.37}} & \textbf{\underline{0.86}} & \textbf{\underline{0.76}} & \textbf{\underline{0.78}} & \textbf{\underline{0.71}} \\
  \bottomrule
\end{tabular}
}
\caption{Impact of ranking information from our prompt ranking method across five datasets. None: not providing ranking information explicitly, Random: providing the randomized order of the three queries as ranking information, and results for this method are averaged over five runs (seeds 0 to 4), ACC: accuracy, F1: Macro-F1 score, \textbf{Bold}: the highest score for each model, \textbf{\underline{Bold}}: the highest score across all models for each dataset.}

\label{tab:prompt_ranking_ablation}
\end{table*}

\subsection{Impact of Ranking Information} 
To investigate the factors behind the success of our prompt ranking method, we conduct experiments to determine whether providing the ranking information is essential for performance gains. In the None setting, all three queries---Counterargument, Explanation, and Goal---are presented without any ranking, meaning the ranking information is not provided to the model. In the Random setting, the queries are randomly shuffled, disregarding their confidence scores, and this randomized ranking information is provided.

Table~\ref{tab:prompt_ranking_ablation} shows that Prompt Ranking consistently outperforms both Random and None settings, underscoring the importance of confidence-based ranking in improving performance. For example, in the CLIMATE dataset, \texttt{gpt-4} with prompt ranking achieves an accuracy of 0.78 and an F1 score of 0.71, while Random Ranking results in a significant drop in the F1 score, highlighting the value of explicitly including the ranking information.

\begin{figure}[!t]
\centering

\includegraphics[width=\linewidth]{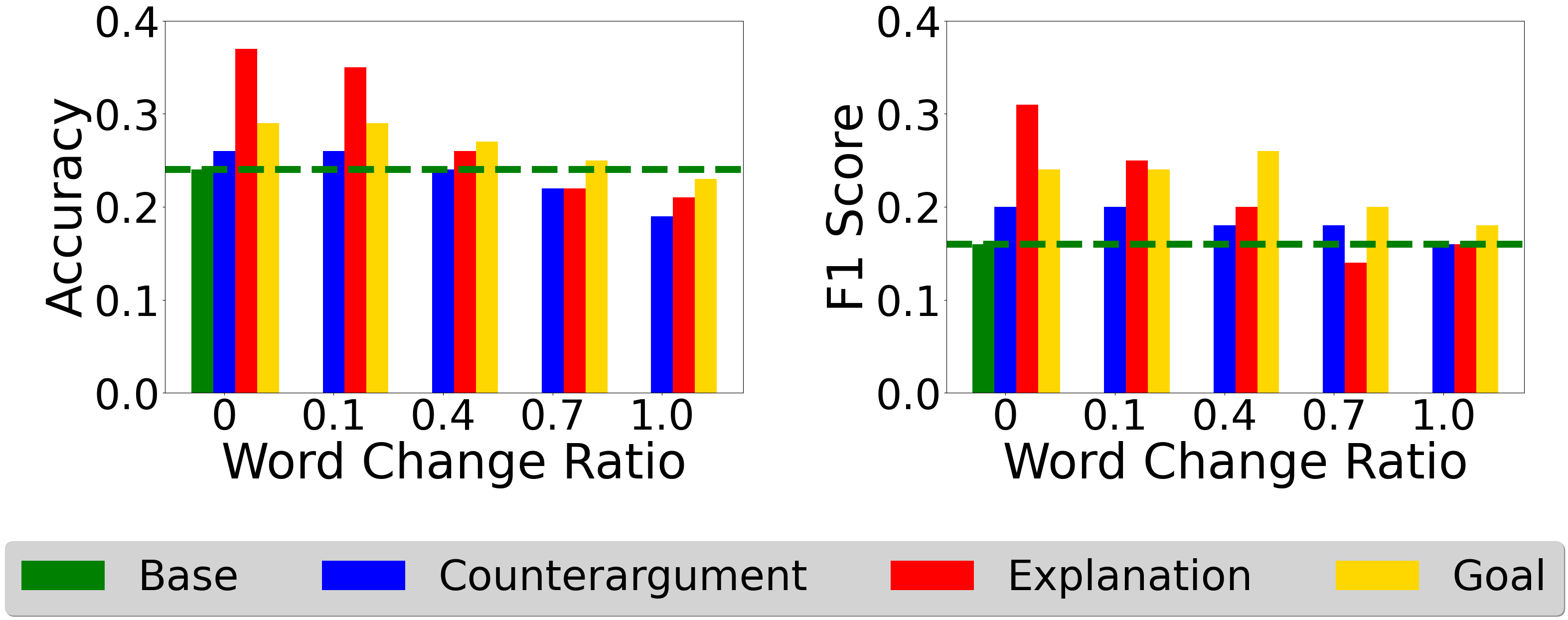}
\caption{Impact of word change ratio on accuracy and Macro-F1 score for each query type.} 

\label{fig:ratio}

\end{figure}

\begin{table}[!t]
\centering
\tiny
\resizebox{\columnwidth}{!}{
\begin{tabular}{lcccc}
\toprule
\textbf{Method} & \textbf{Acc} & \textbf{P} & \textbf{R} & \textbf{F1} \\
\midrule
\textit{Base} & 0.12 & 0.20 & 0.13 & 0.07 \\
\rowcolor{gray!20}\multicolumn{5}{c}{Contextual Augmentations~\cite{sourati2023case}} \\
\textit{+ Counterargument} & 0.13 & 0.12 & 0.11 & 0.08 \\
\textit{+ Explanation}& 0.23 & 0.14 & 0.18 & 0.11 \\
\textit{+ Goal}& 0.19 & 0.16 & 0.17 & 0.11 \\
\rowcolor{gray!20}\multicolumn{5}{c}{Reformulated Query based on \citet{sourati2023case}} \\
\textit{+ Counterargument} & 0.12 & 0.11 & 0.13 & 0.07 \\
\textit{+ Explanation}& 0.21 & 0.15 & 0.14 & 0.09 \\
\textit{+ Goal}& 0.16 & 0.13 & 0.12 & 0.08 \\
\rowcolor{gray!20}\multicolumn{5}{c}{Contextual Augmentations (Ours)} \\
\textit{+ Counterargument} & 0.19 & 0.15 & 0.12 & 0.11 \\
\textit{+ Explanation}& 0.20 & 0.13 & 0.15 & 0.10 \\
\textit{+ Goal}& 0.28 & 0.16 & 0.18 & 0.14 \\
\rowcolor{gray!20}\multicolumn{5}{c}{Reformulated Query from (Ours)} \\
\textit{+ Counterargument} & 0.24 & 0.19 & 0.16 & 0.12 \\
\textit{+ Explanation}& \textbf{0.33} & 0.17 & 0.19 & 0.15 \\
\textit{+ Goal}& \textbf{0.33} & \textbf{0.20} & \textbf{0.20} & \textbf{0.16} \\
\bottomrule
\end{tabular}
}
\caption{Performance comparison based on different contextual augmentation prompt methods and their resulting queries on the PROPAGANDA dataset using the \texttt{gpt-3.5-turbo} model. Acc: Accuracy, P: precision, R: recall, and F1: Macro-F1 score.}

\label{tab:propaganda_ablation}

\end{table}

\begin{table*}[!t]
    \centering 
    \includegraphics[width=\textwidth]{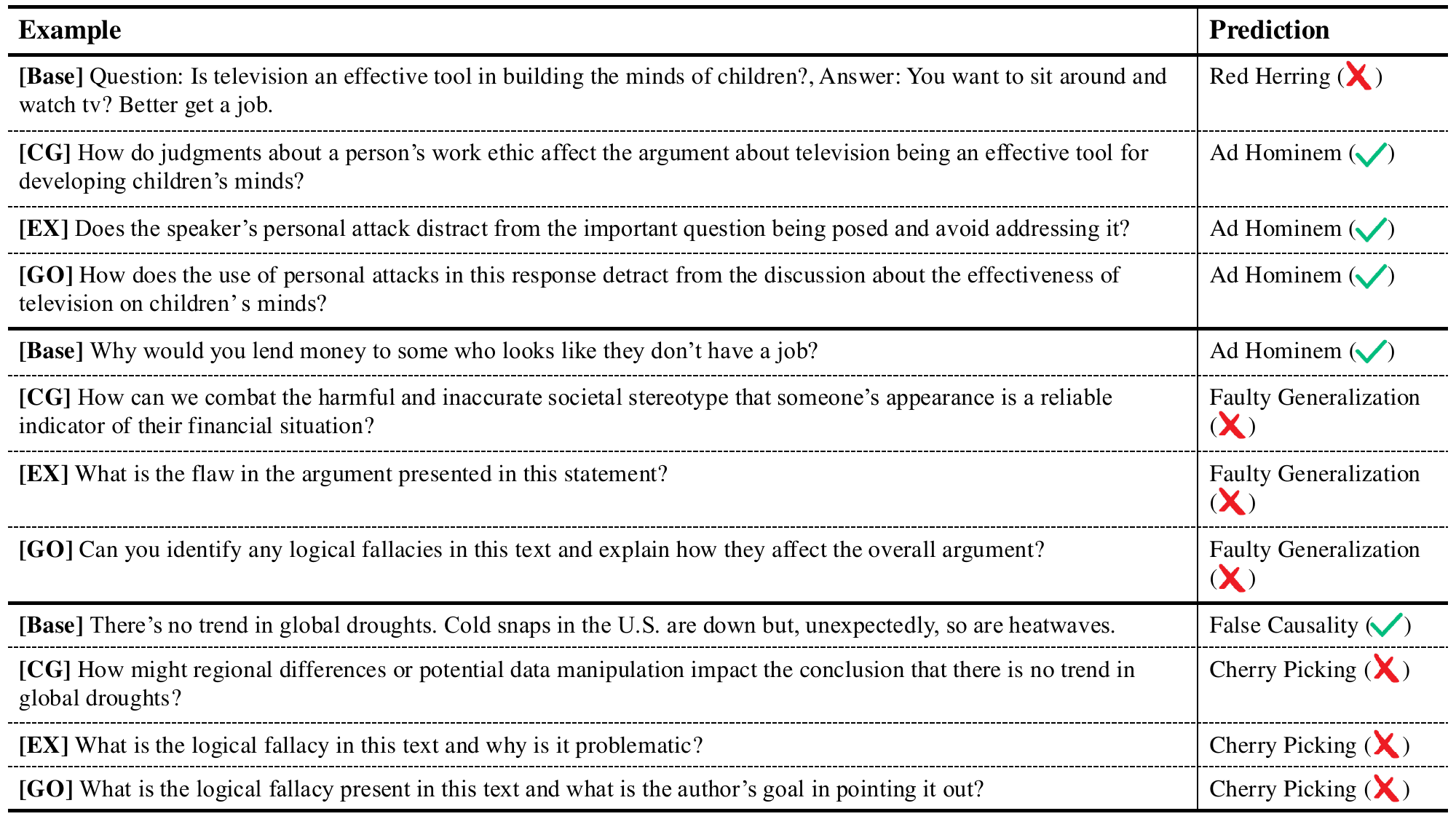}
    \caption{Illustrative examples of query impact on \texttt{gpt-4} performance. Base: the method without using any queries, CG: Counterargument, EX: Explanation, GO: Goal.}
    
\label{tab:analysis}
\end{table*}

\subsection{Impact of Query Structure}\label{validation_of_query}

This analysis aims to determine whether the performance gains of our model in fallacy detection come from the specific structure of our queries (i.e., counterargument, explanation, and goal) or simply from adding extra information, regardless of accuracy or relevance. To test this, we compare our structured queries to randomly generated ones. We hypothesize that if our queries significantly outperform random ones, it would confirm that their structure is key to the observed performance gains. By replacing words in the reformulated queries with arbitrarily similar words, we assess the impact on logical reasoning. 

For the analysis, we randomly select 100 samples from the validation sets of five datasets. To maximize diversity, we repeat the randomization process five times and choose the set with the highest number of unique fallacy classes, resulting in 100 samples representing 18 distinct fallacy types.

To generate random queries, we replace content words in the structured queries with semantically related words obtained from ConceptNet  \cite{speer2017conceptnet}, by selecting the nearest neighbors of each target word. This approach generates queries that are topically relevant but lack the counterargument, explanation, and goal structure. We opt for topically relevant replacements rather than entirely random words to ensure a fair comparison. Although other resources like WordNet \cite{miller1995wordnet} could be used, ConceptNet is chosen for its broader range of semantic relationships, though we believe the choice of resource would not significantly affect the analysis. Figure~\ref{fig:ratio} shows that performance declines as the word change rate increases. This demonstrates that the effect of our method relies on structured logical reasoning, not merely the addition of some information.

\subsection{Impact of Our Approach for Leveraging Implicit Information}\label{compare sourati}

To investigate the effect of our approach for leveraging implicit information, we compare it to the method used by \citet{sourati2023case}, which also incorporates implicit information. The details of the previous method are provided in Appendix~\ref{appendix:zcot,def}. We generate implicit information using both our prompts and theirs and compare the performance of using the information generated by each. We use the \texttt{gpt-3.5-turbo-instruct} model to generate contextual augmentations and queries. 

As shown in Table~\ref{tab:propaganda_ablation}, our queries outperform theirs on the PROPAGANDA dataset. For example, when using our queries based on Explanation and Goal, the classification accuracy is 0.33, compared to 0.23 when using their method. The same trend is observed across other datasets, confirming that our approach for generating contextual augmentations and queries consistently outperforms theirs in detecting logical fallacies.

\subsection{Illustrative Examples of Query Impact}

Table~\ref{tab:analysis} provides examples illustrating how different query types (Counterargument (CG), Explanation (EX), Goal (GO)) affect the model's performance compared to the Base method, which does not use any queries.

The first example illustrates a situation where the Base method fails to detect the fallacy, but all three query types successfully guide the model to the correct classification. Specifically, CG focuses on evaluating judgments about the speaker’s work ethic, EX examines how personal attacks shift attention away from the main argument, and GO highlights how these personal attacks detract from the discussion of television effectiveness on children's minds. This shows how queries can help the model focus on the relevant reasoning behind the argument and accurately identify the fallacy.

In contrast, the next two examples highlight cases where the Base method correctly identifies logical fallacies but fails when queries are applied. In these cases, both EX and GO queries only confirm the presence of a fallacy without challenging the logic behind it, while the CG queries introduce unrelated topics, such as societal stereotypes or regional differences, further distracting the model from identifying the correct fallacy.

These findings suggest that queries can hinder and improve model performance, depending on how well queries align with the intricate logical structure of the argument. When queries fail to direct the model to analyze the core logic or introduce extraneous information, as seen with CG queries in the second and third examples—they can impair performance. Conversely, when queries focus the model on critical aspects of the argument, as in the first example, they can improve performance. Therefore, the effectiveness of each query type depends on its alignment with the underlying logic of the argument, which merits further exploration to determine whether this can be achieved through zero-shot prompting in future work.

\section{Conclusion}\label{conclusion} 
\vspace{-1mm}
In this paper, we presented a simple yet powerful prompt formulation approach for logical fallacy detection. We demonstrated the value of incorporating additional contextual information structured as counterarguments, explanations, and goals, combined with prompt ranking based on confidence scores. Our extensive experiments consistently showed improvements over baselines and state-of-the-art models in both unsupervised and supervised settings, highlighting the robustness and adaptability of our approach.


\section{Limitations} 

While this study demonstrates promising advancements in leveraging Large Language Models (LLMs) like GPT-4 for detecting and classifying logical fallacies through a novel prompt formulation approach, we acknowledge several limitations.

First, the five datasets used in our evaluation may not fully capture the diversity and complexity of real-world scenarios, such as logical reasoning in fields like medicine and science. Additionally, there are fallacies beyond the 29 types we explored. For instance, the type-token fallacy, which occurs when a word can refer to either a type (an abstract descriptive concept) or a token (an object that instantiates a concept) and is used ambiguously, was not included in our analysis.

Furthermore, our method's reliance on specific models (\texttt{gpt-3.5-turbo}, \texttt{gpt-4}, \texttt{Llama2-7b-hf}, \texttt{Llama2-13b-hf}, and \texttt{Llama3-8B}) may limit its applicability to other LLMs or future iterations, potentially impacting scalability and cost-effectiveness due to the computational resources required for prompt reformulation. 

In addition, while the improvements in Macro-F1 scores are significant, further research is needed to enhance the interpretability and transparency of the reformulation process to ensure broader applicability in real-world settings.

Lastly, variations in query performance across fallacy classes may stem from differences in dataset-specific definitions and class imbalance. For instance, in the datasets for our experiments, more prevalent classes like Loaded Language allowed for clearer evaluations, while underrepresented classes like Strawman limited meaningful comparisons. This underscores the need to address class imbalance to ensure more consistent evaluations in future studies.

\section{Ethics Statement}

Our methodology aims to enhance logical fallacy detection by employing contextual augmentations and structured prompt reformulation. We do not utilize any external knowledge or information that might bias the evaluation of our model. However, while our approach improves model performance in identifying logical fallacies, it may inadvertently amplify existing biases present in the training data of LLMs. We acknowledge this limitation and recommend further research to mitigate potential biases in logical reasoning tasks.

\section{Acknowledgements}
This work was supported by the Institute of Information \& Communications Technology Planning \& evaluation (IITP) grant, and the National Research Foundation of Korea (NRF) grant funded by the Korean government (MSIT) (RS-2019-II190421, IITP-2025-RS-2020-II201821, RS-2024-00438686, RS-2024-00436936, RS-2024-00360227, RS-2023-0022544, NRF-2021M3H4A1A02056037, RS-2024-00448809).  This research was also partially supported by the Culture, Sports, and Tourism R\&D Program through the Korea Creative Content Agency grant funded by the Ministry of Culture, Sports and Tourism in 2024 (RS-2024-00333068). Further support was provided by Indiana University Indianapolis.

\bibliography{custom}

\newpage

\newpage
\appendix

\section{Overview}\label{appendix:overview}
In this appendix, we provide detailed explanations and additional results related to the methods used in the main study. Section~\ref{appendix:prompt} introduces the prompt techniques, including ZCoT, DEF, and the contextual augmentation method, along with examples of our four-step prompt technique. Section~\ref{appendix:exp} outlines the grouping of similar fallacy classes across datasets, as well as the dataset splits for supervised and unsupervised settings.  Section~\ref{appendix:classification} presents multi-class classification results for \modelLLaMA models. Finally, Section~\ref{appendix:calibration} provides additional information on calibration results across five datasets for Base, Ours (Prompt Ranking), ZCoT~\cite{kojima2022large}, DEF~\cite{hong2024closer}, and individual queries.

\section{Prompt Details}\label{appendix:prompt}

\subsection{Prompt Techniques: ZCoT, DEF, and Contextual Augmentations}\label{appendix:zcot,def}
Table~\ref{tab:prompt-baseline} compares the prompt methods used for multi-class classification. The ZCoT \cite{kojima2022large} method adds the phrase "Let's think step by step" to guide inference, applying a straightforward prompt that detects labels directly in multi-class settings. In the original DEF \cite{hong2024closer} method, when a fallacy sentence is given, the model is provided with the fallacy class to which the sentence belongs, along with the definition of that class, and is then tasked with determining whether the sentence contains a fallacy. This approach is designed for binary classification. However, for multi-class classification, providing all fallacy classes and definitions upfront can result in a form of cheating, so it cannot be directly applied. To address this, we modify the prompt for multi-class classification by providing all the logical fallacy classes in the dataset along with their definitions, instructing the model to classify the type of fallacy present in the sentence.

Additionally, \citet{sourati2023case}'s approach uses ChatGPT to generate multiple variations of each contextual augmentation (e.g., counterargument, explanation, and goal), which are manually revised, while our fully automated method employs \texttt{gpt-3.5-turbo-instruct} to generate a single output without manual adjustment. Their prompts solely focus on these three perspectives (counterargument, explanation, goal), whereas ours not only incorporates these perspectives but also explicitly includes fallacy classes, leading to more effective logical fallacy detection (see Table~\ref{tab:prompt}).

\subsection{Example about Our Four-Step Prompts}
\label{appendix:four_step}

\subsubsection{Contextual Augmentation Prompts}
\label{appendix1}
These prompts are utilized to generate counterarguments, explanations, and goals directly from the LLM, forming the contextual augmentation texts (\(R_i(x)\)) as outlined in Equation~\ref{equation1}. This foundational step, represented by \(\mathcal{R}_i(x) = \text{LLM}(x, \mathcal{I}^i)\), allows us to delve deeper into the logical structure and validity of the argument presented in the text. Here, \(x\) represents the input text and \(\mathcal{I}^i\) signifies the instruction for generating the counterargument (\(\mathcal{I}^1\)), explanation (\(\mathcal{I}^2\)), or goal (\(\mathcal{I}^3\)) of \(x\). Fallacy classes refer to the labels assigned to each dataset that categorize the specific types of logical fallacies. Consider the [TEXT]: \textit{"Annie must like Starbucks because all girls like Starbucks."}

\begin{itemize}
    \item \textbf{Counterargument} \\
    $\blacktriangleright$ \textbf{I'll give you some texts. The texts can be question and answer pairs or sentences. The text contains one of the following logical fallacies:\{fallacy classes\}. Represent the counterargument to the text.} \\
    $\blacktriangleright$ \textbf{Text:} [TEXT] \\
    $\blacktriangleright$ \textbf{Counterargument (Example):} \textit{"Not all girls like Starbucks, as personal preferences vary among individuals. Even if Annie is a girl, it does not automatically mean that she likes Starbucks. She may prefer a different type of coffee or may not like coffee at all. It is not fair to make assumptions about someone based on their gender."}
    \item \textbf{Explanation} \\ 
    $\blacktriangleright$ \textbf{I'll give you some texts. The texts can be question and answer pairs or sentences. The text contains one of the following logical fallacies:\{fallacy classes\}. Analyze the text.} \\
    $\blacktriangleright$ \textbf{Text:} [TEXT] \\
    $\blacktriangleright$ \textbf{Explanation (Example):} \textit{"This text suggests a generalization about girls and their preferences for Starbucks, assuming that Annie, as a girl, must also like Starbucks without evidence. This could be seen as stereotyping, making unfounded assumptions based on gender, reinforcing harmful stereotypes."}
    \item \textbf{Goal} \\
    $\blacktriangleright$ \textbf{I'll give you some texts. The texts can be question and answer pairs or sentences. The text contains one of the following logical fallacies:\{fallacy classes\}. Express the goal of the text.} \\
    $\blacktriangleright$ \textbf{Text:} [TEXT] \\
    $\blacktriangleright$ \textbf{Goal (Example):} \textit{"The goal is to make a generalization about girls liking Starbucks based on the assumption that Annie is a girl."}
\end{itemize}

\subsubsection{Reformulated Query Prompts}
\label{appendix2}
Following the generation of augmented texts, these prompts are employed to craft queries that delve deeper into the logical structure and validity of the arguments presented in the texts, as denoted by the equation \(\quad \mathcal{Q}_{i} = \text{LLM}(x, \mathcal{R}_i)\). In this equation, \(\mathcal{R}_i\) represents the contextual augmentation obtained from the input text \(x\), These generated queries are intended to enhance the depth of analysis of \(x\), improving the detection of logical fallacies. The indices \(i = 1, 2, 3\) correspond to the different aspects of queries: Counterargument (\(\mathcal{Q}_1\)), Explanation (\(\mathcal{Q}_2\)), and Goal (\(\mathcal{Q}_3\)). Below are the detailed prompts for generating these queries based on the augmented texts:

\begin{itemize}
    \item \textbf{Query Generation for Counterargument Text}\\
    $\blacktriangleright$ \textbf{I'll give you some texts and their counterarguments. The texts can be question and answer pairs or sentences. Create one query for each text to analyze the text based on its counterarguments.} \\
    $\blacktriangleright$ \textbf{Text:} [TEXT] \\
    $\blacktriangleright$ \textbf{Counterargument:} [COUNTERARGUMENT] \\
    $\blacktriangleright$ \textbf{Query (Example):} \textit{"How does the counterargument challenge the assumption that all girls like Starbucks?"}
    \item \textbf{Query Generation for Explanation Text} \\ 
    $\blacktriangleright$ \textbf{I'll give you some texts and their explanations. The texts can be question and answer pairs or sentences. Create one query for each text to analyze the text based on its explanations.} \\
    $\blacktriangleright$ \textbf{Text:} [TEXT] \\
    $\blacktriangleright$ \textbf{Explanation:} [EXPLANATION] \\
    $\blacktriangleright$ \textbf{Query (Example):} \textit{"How does this text perpetuate harmful gender stereotypes and restrict individual expression?"}
    \item \textbf{Query Generation for Goal Text} \\
    $\blacktriangleright$ \textbf{I'll give you some texts and their goals. The texts can be question and answer pairs or sentences. Create one query for each text to analyze the text based on its goal.} \\
    $\blacktriangleright$ \textbf{Text:} [TEXT] \\
    $\blacktriangleright$ \textbf{Goal:} [GOAL] \\
    $\blacktriangleright$ \textbf{Query (Example):} \textit{"What does this text reveal about the speaker's attitude towards girls and their preferences?"}
\end{itemize}

\subsubsection{Confidence Score Calculation for Queries}
\label{appendix3}
In our experiments aimed at classifying logical fallacies, we calculate the confidence scores for each query as described in Step~\ref{step3}. This process evaluates the probability \(p_{\text{LLM}}\) of a fallacy's class label (\(l\)) for a given input text \(x\) based on the generated query \(\mathcal{Q}_i\). The confidence score for each query is computed by summing the log probabilities of the predicted tokens, which indicates how confidently the model predicts the most probable class based on the query type. These scores are then used for ranking and classification in subsequent steps.

Each \textbf{formulated prompt} involves one of three query types: Counterargument, Explanation, and Goal. Below, we provide a detailed explanation of the prompts used in this analysis:

\begin{itemize}
    \item \textbf{Logical Fallacy Multi-Class Classification} \\
    $\blacktriangleright$ \textbf{Your task is to classify the type of fallacy in the Text. The label can be `Appeal to Emotion', `Faulty Generalization', `Red Herring', `Ad Hominem', and `Irrelevant Authority'. Please classify the type of fallacy in the Text based on the Query.} \\
    $\blacktriangleright$ \textbf{Text:} [TEXT] \\
    $\blacktriangleright$ \textbf{Formulated Prompt:}  [ONE OF THREE QUERIES]\\
    $\blacktriangleright$ \textbf{Label:}
\end{itemize}

\begin{table*}[!ht]
\centering
\resizebox{\textwidth}{!}{
\begin{tabular}{>{\centering\arraybackslash}m{3cm} lllllllllllll}
\toprule
\textbf{LLM} & \textbf{Method} & \multicolumn{2}{c}{\textbf{PROPAGANDA}} & \multicolumn{2}{c}{\textbf{ARGOTARIO}} & \multicolumn{2}{c}{\textbf{LOGIC}} & \multicolumn{2}{c}{\textbf{COVID-19}} & \multicolumn{2}{c}{\textbf{CLIMATE}} \\ \cmidrule(lr){3-4} \cmidrule(lr){5-6} \cmidrule(lr){7-8} \cmidrule(lr){9-10} \cmidrule(lr){11-12}
 &  & \multicolumn{1}{c}{\textbf{ACC}} & \multicolumn{1}{c}{\textbf{F1}} & \multicolumn{1}{c}{\textbf{ACC}} & \multicolumn{1}{c}{\textbf{F1}} & \multicolumn{1}{c}{\textbf{ACC}} & \multicolumn{1}{c}{\textbf{F1}} & \multicolumn{1}{c}{\textbf{ACC}} & \multicolumn{1}{c}{\textbf{F1}} & \multicolumn{1}{c}{\textbf{ACC}} & \multicolumn{1}{c}{\textbf{F1}} \\ \midrule \midrule
\multirow{6}{*}{\texttt{Llama2-7b-hf}} & Zero-shot & \multicolumn{1}{c}{0.08} & \multicolumn{1}{c}{0.04} & \multicolumn{1}{c}{0.20} & \multicolumn{1}{c}{0.16} & \multicolumn{1}{c}{0.14} & \multicolumn{1}{c}{0.07} & \multicolumn{1}{c}{0.07} & \multicolumn{1}{c}{0.05} & \multicolumn{1}{c}{0.23} & \multicolumn{1}{c}{0.10} \\
 & ZCoT \cite{kojima2022large} &  \multicolumn{1}{c}{0.10} & \multicolumn{1}{c}{0.05} & \multicolumn{1}{c}{0.15} & \multicolumn{1}{c}{0.12} & \multicolumn{1}{c}{0.15} & \multicolumn{1}{c}{0.07} & \multicolumn{1}{c}{0.14} & \multicolumn{1}{c}{0.06} & \multicolumn{1}{c}{0.13} & \multicolumn{1}{c}{0.03} \\
 & DEF \cite{hong2024closer} & \multicolumn{1}{c}{0.10} & \multicolumn{1}{c}{0.05} & \multicolumn{1}{c}{0.26} & \multicolumn{1}{c}{0.08} & \multicolumn{1}{c}{0.15} & \multicolumn{1}{c}{0.07} & \multicolumn{1}{c}{0.14} & \multicolumn{1}{c}{0.06} & \multicolumn{1}{c}{0.13} & \multicolumn{1}{c}{0.03} \\
 & \textit{Counterargument} & \multicolumn{1}{c}{0.10} & \multicolumn{1}{c}{0.05} & \multicolumn{1}{c}{0.25} & \multicolumn{1}{c}{0.20} & \multicolumn{1}{c}{0.20} & \multicolumn{1}{c}{0.16} & \multicolumn{1}{c}{0.21} & \multicolumn{1}{c}{0.14} & \multicolumn{1}{c}{0.14} & \multicolumn{1}{c}{0.10} \\
 & \textit{Explanation} & \multicolumn{1}{c}{0.15} & \multicolumn{1}{c}{\textbf{0.08}} & \multicolumn{1}{c}{\textbf{0.36}} & \multicolumn{1}{c}{\textbf{0.29}} & \multicolumn{1}{c}{\textbf{0.37}} & \multicolumn{1}{c}{\textbf{0.32}} & \multicolumn{1}{c}{\textbf{0.29}} & \multicolumn{1}{c}{\textbf{0.24}} & \multicolumn{1}{c}{\textbf{\underline{0.41}}} & \multicolumn{1}{c}{\textbf{0.32}} \\
 & \textit{Goal} & \multicolumn{1}{c}{0.13} & \multicolumn{1}{c}{0.07} & \multicolumn{1}{c}{0.35} & \multicolumn{1}{c}{\textbf{0.29}} & \multicolumn{1}{c}{0.31} & \multicolumn{1}{c}{0.26} & \multicolumn{1}{c}{\textbf{0.29}} & \multicolumn{1}{c}{0.16} & \multicolumn{1}{c}{0.34} & \multicolumn{1}{c}{0.24} \\
 & Prompt Ranking (Ours) & \multicolumn{1}{c}{\textbf{0.21}} & \multicolumn{1}{c}{0.06} & \multicolumn{1}{c}{0.26} & \multicolumn{1}{c}{0.20} & \multicolumn{1}{c}{0.23} & \multicolumn{1}{c}{0.19} & \multicolumn{1}{c}{\textbf{0.29}} & \multicolumn{1}{c}{0.13} & \multicolumn{1}{c}{0.24} & \multicolumn{1}{c}{0.16} \\
  \midrule
\multirow{6}{*}{\texttt{Llama2-13b-hf}} & Zero-shot & \multicolumn{1}{c}{0.08} & \multicolumn{1}{c}{0.03} & \multicolumn{1}{c}{0.22} & \multicolumn{1}{c}{0.18} & \multicolumn{1}{c}{0.16} & \multicolumn{1}{c}{0.08} & \multicolumn{1}{c}{0.14} & \multicolumn{1}{c}{0.03} & \multicolumn{1}{c}{0.17} & \multicolumn{1}{c}{0.05} \\
 & ZCoT \cite{kojima2022large} &  \multicolumn{1}{c}{0.13} & \multicolumn{1}{c}{0.05} & \multicolumn{1}{c}{0.22} & \multicolumn{1}{c}{0.18} & \multicolumn{1}{c}{0.16} & \multicolumn{1}{c}{0.08} & \multicolumn{1}{c}{0.07} & \multicolumn{1}{c}{0.05} & \multicolumn{1}{c}{0.14} & \multicolumn{1}{c}{0.05} \\
 & DEF \cite{hong2024closer} & \multicolumn{1}{c}{\textbf{0.29}} & \multicolumn{1}{c}{0.05} & \multicolumn{1}{c}{0.27} & \multicolumn{1}{c}{0.17} & \multicolumn{1}{c}{0.18} & \multicolumn{1}{c}{0.05} & \multicolumn{1}{c}{0.07} & \multicolumn{1}{c}{0.02} & \multicolumn{1}{c}{0.24} & \multicolumn{1}{c}{0.08} \\
 & \textit{Counterargument} & \multicolumn{1}{c}{0.14} & \multicolumn{1}{c}{0.07} & \multicolumn{1}{c}{0.25} & \multicolumn{1}{c}{0.19} & \multicolumn{1}{c}{0.21} & \multicolumn{1}{c}{0.14} & \multicolumn{1}{c}{0.29} & \multicolumn{1}{c}{0.15} & \multicolumn{1}{c}{0.21} & \multicolumn{1}{c}{0.10} \\
 & \textit{Explanation} & \multicolumn{1}{c}{0.22} & \multicolumn{1}{c}{\textbf{\underline{0.10}}} & \multicolumn{1}{c}{\textbf{0.43}} & \multicolumn{1}{c}{\textbf{0.34}} & \multicolumn{1}{c}{\textbf{0.38}} & \multicolumn{1}{c}{\textbf{0.32}} & \multicolumn{1}{c}{0.21} & \multicolumn{1}{c}{0.11} & \multicolumn{1}{c}{\textbf{0.35}} & \multicolumn{1}{c}{\textbf{0.26}} \\
 & \textit{Goal} & \multicolumn{1}{c}{0.20} & \multicolumn{1}{c}{0.08} & \multicolumn{1}{c}{\textbf{0.40}} & \multicolumn{1}{c}{\textbf{0.32}} & \multicolumn{1}{c}{0.37} & \multicolumn{1}{c}{0.27} & \multicolumn{1}{c}{\textbf{\underline{0.36}}} & \multicolumn{1}{c}{\textbf{\underline{0.29}}} & \multicolumn{1}{c}{0.28} & \multicolumn{1}{c}{0.20} \\
 & Prompt Ranking (Ours) & \multicolumn{1}{c}{\textbf{\underline{0.34}}} & \multicolumn{1}{c}{0.08} & \multicolumn{1}{c}{0.27} & \multicolumn{1}{c}{0.20} & \multicolumn{1}{c}{0.26} & \multicolumn{1}{c}{0.21} & \multicolumn{1}{c}{0.14} & \multicolumn{1}{c}{0.03} & \multicolumn{1}{c}{0.29} & \multicolumn{1}{c}{0.19} \\
  \midrule
\multirow{6}{*}{\texttt{Llama3-8B}} & Zero-shot & \multicolumn{1}{c}{0.13} & \multicolumn{1}{c}{0.05} & \multicolumn{1}{c}{0.21} & \multicolumn{1}{c}{0.13} & \multicolumn{1}{c}{0.25} & \multicolumn{1}{c}{0.13} & \multicolumn{1}{c}{0.07} & \multicolumn{1}{c}{0.02} & \multicolumn{1}{c}{0.14} & \multicolumn{1}{c}{0.08} \\
 & ZCoT \cite{kojima2022large} &  \multicolumn{1}{c}{0.23} & \multicolumn{1}{c}{0.04} & \multicolumn{1}{c}{0.19} & \multicolumn{1}{c}{0.13} & \multicolumn{1}{c}{0.20} & \multicolumn{1}{c}{0.09} & \multicolumn{1}{c}{0.07} & \multicolumn{1}{c}{0.07} & \multicolumn{1}{c}{0.13} & \multicolumn{1}{c}{0.06} \\
 & DEF \cite{hong2024closer} & \multicolumn{1}{c}{\textbf{0.28}} & \multicolumn{1}{c}{0.06} & \multicolumn{1}{c}{0.25} & \multicolumn{1}{c}{0.13} & \multicolumn{1}{c}{0.23} & \multicolumn{1}{c}{0.11} & \multicolumn{1}{c}{0.07} & \multicolumn{1}{c}{0.03} & \multicolumn{1}{c}{0.14} & \multicolumn{1}{c}{0.07} \\
 & \textit{Counterargument} & \multicolumn{1}{c}{0.13} & \multicolumn{1}{c}{0.06} & \multicolumn{1}{c}{0.34} & \multicolumn{1}{c}{0.26} & \multicolumn{1}{c}{0.29} & \multicolumn{1}{c}{0.20} & \multicolumn{1}{c}{0.21} & \multicolumn{1}{c}{0.21} & \multicolumn{1}{c}{0.15} & \multicolumn{1}{c}{0.11} \\
 & \textit{Explanation} & \multicolumn{1}{c}{0.20} & \multicolumn{1}{c}{\textbf{0.09}} & \multicolumn{1}{c}{0.51} & \multicolumn{1}{c}{\textbf{\underline{0.42}}} & \multicolumn{1}{c}{\textbf{\underline{0.54}}} & \multicolumn{1}{c}{\textbf{\underline{0.50}}} & \multicolumn{1}{c}{\textbf{\underline{0.36}}} & \multicolumn{1}{c}{0.24} & \multicolumn{1}{c}{\textbf{\underline{0.41}}} & \multicolumn{1}{c}{0.37} \\
 & \textit{Goal} & \multicolumn{1}{c}{0.20} & \multicolumn{1}{c}{\textbf{0.09}} & \multicolumn{1}{c}{\textbf{\underline{0.53}}} & \multicolumn{1}{c}{0.41} & \multicolumn{1}{c}{0.51} & \multicolumn{1}{c}{0.45} & \multicolumn{1}{c}{0.29} & \multicolumn{1}{c}{0.21} & \multicolumn{1}{c}{0.40} & \multicolumn{1}{c}{0.33} \\
 & Prompt Ranking (Ours) & \multicolumn{1}{c}{0.22} & \multicolumn{1}{c}{\textbf{0.09}} & \multicolumn{1}{c}{0.30} & \multicolumn{1}{c}{0.29} & \multicolumn{1}{c}{0.38} & \multicolumn{1}{c}{0.36} & \multicolumn{1}{c}{\textbf{\underline{0.36}}} & \multicolumn{1}{c}{\textbf{0.25}} & \multicolumn{1}{c}{0.33} & \multicolumn{1}{c}{\textbf{\underline{0.40}}} \\
 \bottomrule
\end{tabular}
}
\caption{Accuracy and Macro-F1 scores of the Multi-class fallacy classification on all datasets. This table presents the experimental results for three models in the \modelLLaMA series. \textbf{Bold}: the highest score for each model, \textbf{\underline{bold}}: the highest score across all models for each dataset. \textbf{ZCoT}: Zero-shot Chain of Thought \cite{kojima2022large}, and \textbf{DEF}: the method used by \citet{hong2024closer} for providing definitions of logical fallacies.}
\label{tab:multiclass-llama}
\end{table*}

\subsubsection{Prompt Ranking-Based Classification}
\label{appendix4}

To implement prompt ranking, we append the following instruction to the logical fallacy classification prompt: "Return only the name of the label, and nothing else. MAKE SURE your output is one of the \{n\_{\text{labels}}\}\footnote{\{n\_{\text{labels}}\} represents the number of labels specific to the dataset.}labels stated." This instruction ensures concise outputs, preventing unnecessary information that could reduce the confidence score. We calculate the confidence scores of the results from each generated query by using the log probabilities of the tokens outputted by LLMs (e.g., \texttt{gpt-3.5-turbo}, \texttt{gpt-4}, \texttt{Llama2-7b-hf}, \texttt{Llama2-13b-hf}, and \texttt{Llama3-8B}). These confidence scores are then used to rank queries in descending order, prioritizing those with higher confidence. 

For example, in the Ranking Information provided below, the order of query names is based on their confidence scores: Explanation Query, Goal Query, and Counterargument Query. In this case, the Explanation Query has the highest confidence score, contributing most to the final classification. While the content of each query (\(\mathcal{Q}_i\)) is used in the model's input, the ranking information, which is derived by sorting the confidence scores of these queries, is incorporated into the final prompt, thereby improve the overall reliability of the fallacy classification process.

It is important to note that for \modelLLaMA-based models, the instruction "Return only the name of the label, and nothing else. MAKE SURE your output is one of the \{n\_{\text{labels}}\} labels stated." is not always effectively applied. Therefore, for these models, we extract the log probabilities corresponding to the fallacy class in the generated output and sum these values to calculate the confidence scores.

\begin{itemize}
    \item \textbf{Logical Fallacy Multi-Class Classification} \\
    $\blacktriangleright$ \textbf{Given a sentence with a logical fallacy, we aim to detect it using queries based on multiple perspectives, such as counterargument, explanation, and goal. The ranking information indicates the order of queries based on their confidence scores, which are helpful in identifying the specific type of logical fallacy present in the sentence. The label can be `Appeal to Emotion', `Faulty Generalization', `Red Herring', `Ad Hominem', or `Irrelevant Authority'. Based on the ranking information of these queries, please reference them to detect the fallacy in the sentence.} \\
    $\blacktriangleright$ \textbf{Text:} [TEXT] \\
    $\blacktriangleright$ \textbf{Formulated Prompt:} \\
    \vspace{-4mm}
    \begin{itemize}
        \item \textbf{Counterargument Query}: How does the counterargument challenge the assumption that all girls like Starbucks?
        \item \textbf{Explanation Query}: How does this text perpetuate harmful gender stereotypes and restrict individual expression?
        \item \textbf{Goal Query}: What does this text reveal about the speaker’s attitude towards girls and their preferences?
    \end{itemize}
    $\blacktriangleright$ \textbf{Ranking Information:} Explanation Query, Goal Query, Counterargument Query \\
    $\blacktriangleright$ \textbf{Label:}
\end{itemize}

\section{Dataset Preparation}\label{appendix:exp}

\subsection{Fallacy Class}
To ensure coherence in our analysis, we group similar fallacy classes across the datasets: \textbf{\textit{Hasty Generalization}} and \textbf{\textit{Faulty Generalization}} into \textbf{\textit{Faulty Generalization}}; \textbf{\textit{Fallacy of Credibility}}, \textbf{\textit{False Authority}}, and \textbf{\textit{Appeal to Authority}} into \textbf{\textit{Irrelevant Authority}}; and \textbf{\textit{False Cause}}, \textbf{\textit{False Causality}}, \textbf{\textit{Post Hoc}}, and \textbf{\textit{Causal Oversimplification}} into \textbf{\textit{False Causality}}.

\subsection{Data Description}
For both the supervised and unsupervised settings, the datasets are split into training, validation, and test sets in proportions of 65\%, 15\%, and 20\%, respectively, except for the LOGIC dataset, where we use the predefined test set from \citet{jin2022logical}. The PROPAGANDA dataset, pre-processed by \citet{alhindi2022multitask}, is also shuffled and divided into training, validation, and test sets following the same 65\%, 15\%, and 20\% split.

\section{Multi-Class Classification Results on \modelLLaMA Models}\label{appendix:classification}
Table~\ref{tab:multiclass-llama} presents the results of multi-class fallacy classification using three \modelLLaMA models (\texttt{Llama2-7b-hf\allowbreak}, \texttt{Llama2-13b-hf\allowbreak}, and \texttt{Llama3-8B}
). Unlike the \texttt{gpt-3.5-turbo} and \texttt{gpt-4} results (Table~\ref{tab:multiclass}), the Explanation (EX) query consistently shows the best performance in these models, often surpassing Prompt Ranking (PR). For example, in the \texttt{Llama2-13b-hf} and \texttt{Llama3-8B} models, Explanation scores the highest Macro-F1 and accuracy in datasets like LOGIC and ARGOTARIO. This contrasts with the larger \modelGPT models, where Prompt Ranking leads to the most significant improvements. We hypothesize that this difference arises due to the smaller parameter sizes of the \modelLLaMA models we used, which may limit Prompt Ranking’s effectiveness with complex query combinations. Nonetheless, PR still performs competitively, especially in the PROPAGANDA dataset, demonstrating that ranking-based methods can offer considerable benefits even in smaller models.

\begin{figure}[!b]
\centering

\includegraphics[width=\linewidth]{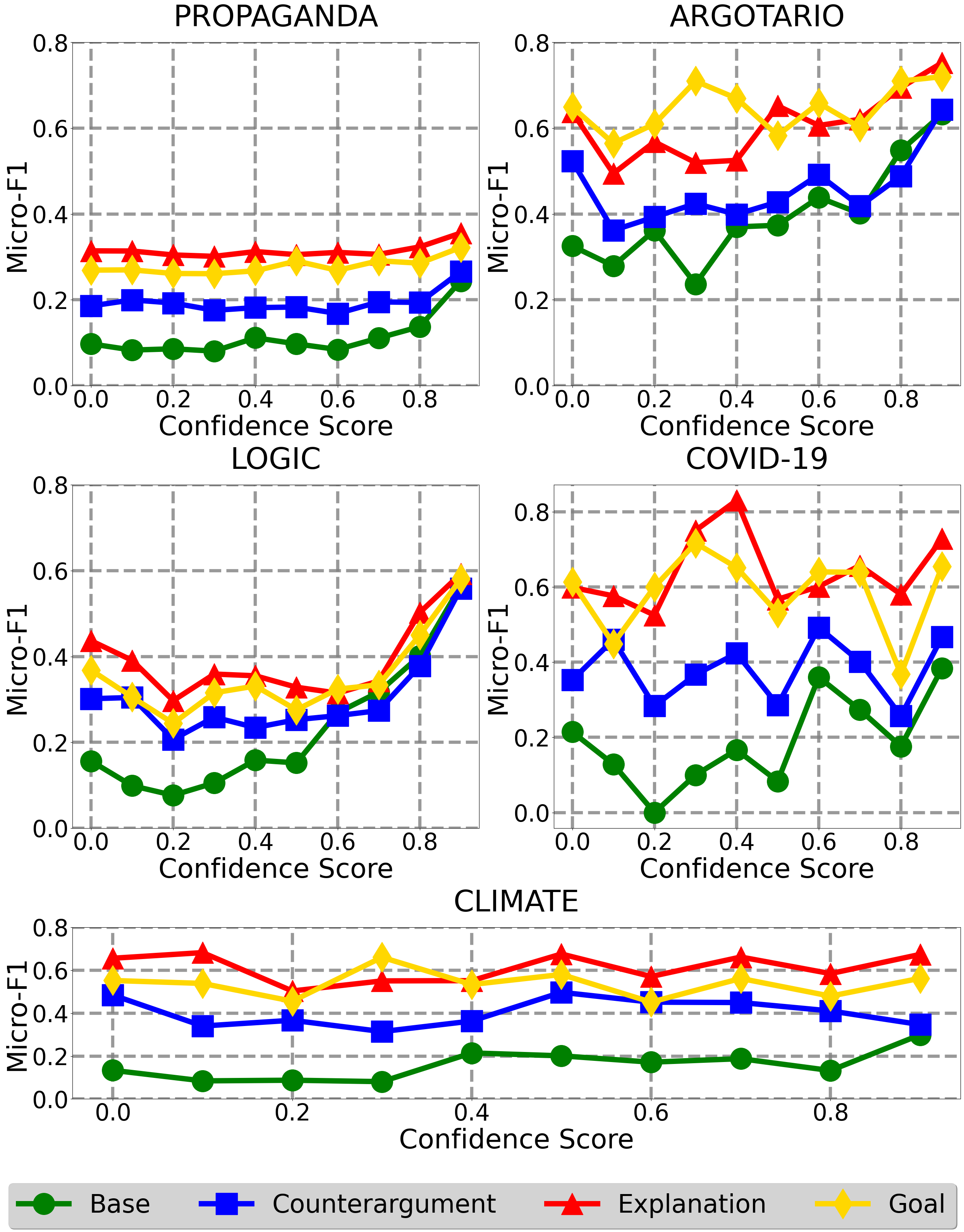}
\caption{Relationship between confidence scores and performance with/without reformulated queries for all datasets.}
\label{fig:confidence_all}
\end{figure}

\section{Additional Calibration Results}\label{appendix:calibration}

As shown in Section~\ref{calibration}, our prompt ranking method (PR) demonstrates better calibration than the Base method, though the individual calibration performance of each query type is not fully detailed. This section provides the calibration results for each query type, offering further insight into their contributions to model reliability.

Across all datasets, PR consistently shows the greatest calibration improvements, reducing the gap between predicted confidence and actual accuracy more effectively than the Base method or individual queries. Among the queries, Explanation (EX) generally performs the best, followed by Goal (GO) and Counterargument (CG), though they do not match the overall improvement achieved by PR.

\begin{figure*}[!ht]
\centering

\includegraphics[width=\linewidth]{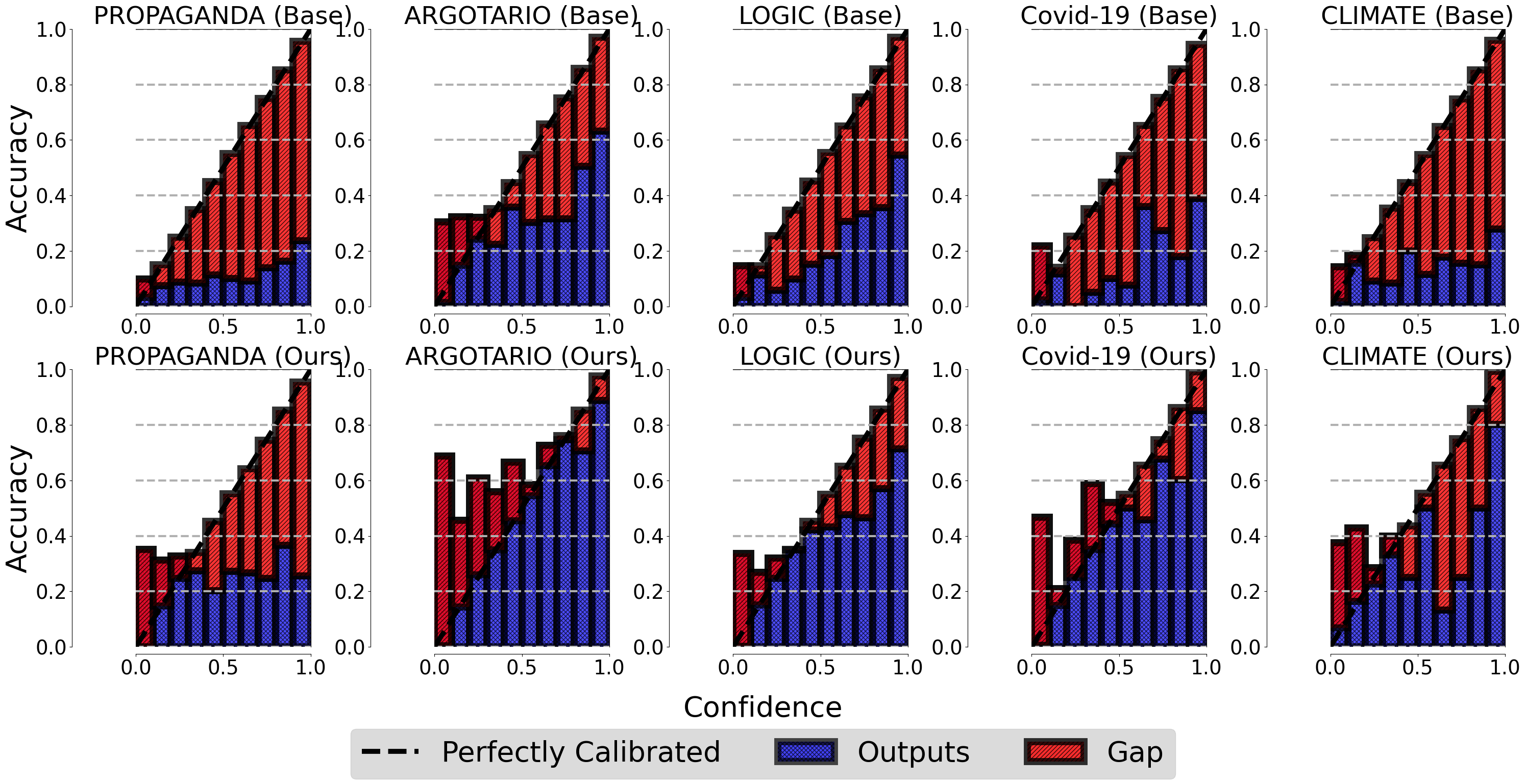}
\caption{Reliability diagrams comparing the calibration performance of the Base method (without reformulated queries) and Ours (Prompt Ranking) across five datasets, both using the \texttt{gpt-3.5-turbo} model.}
\label{fig:calibration_all}
\end{figure*}

\begin{figure*}[!ht]
\centering

\includegraphics[width=0.8\linewidth]{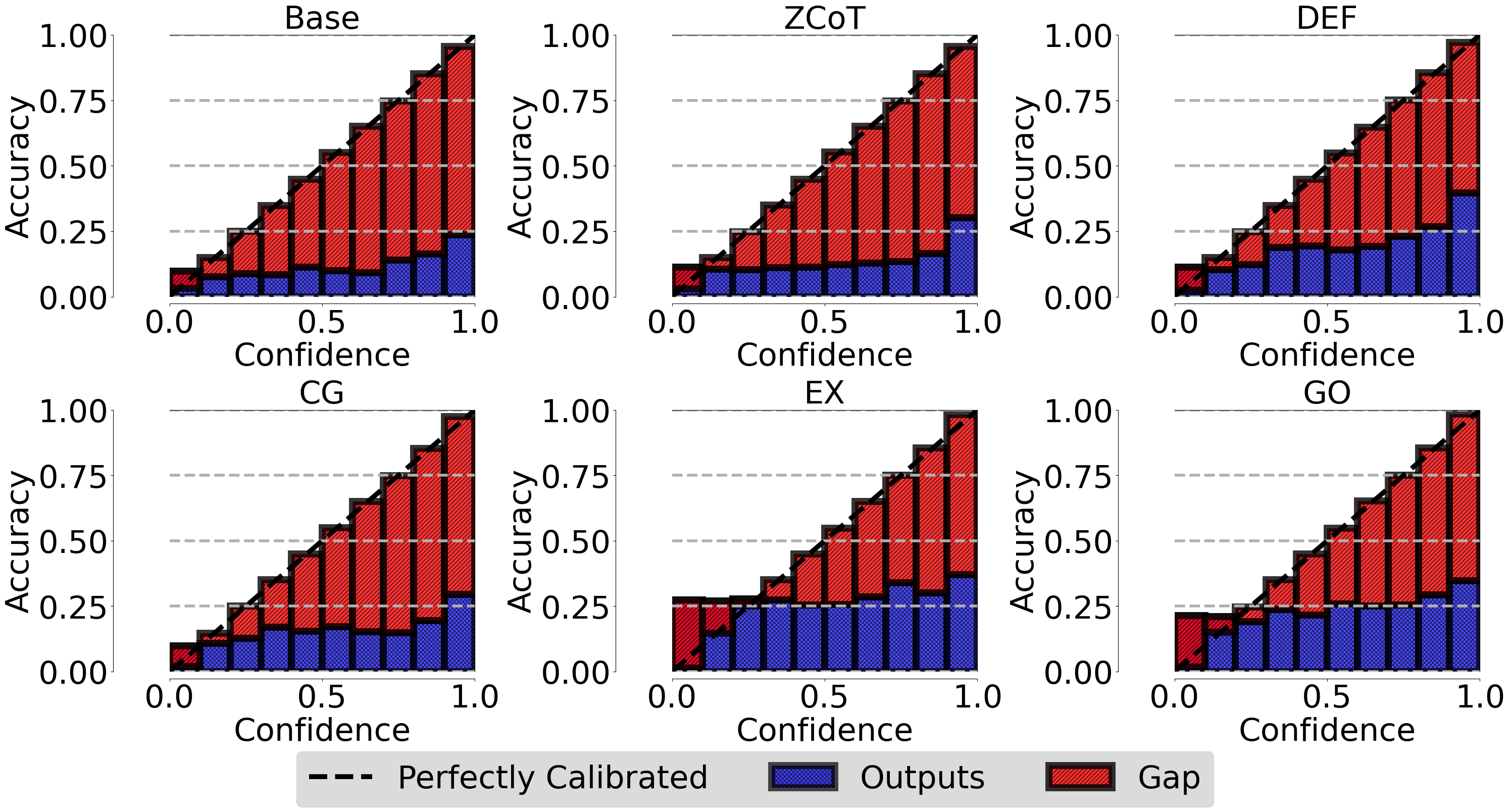}
\caption{Reliability diagrams comparing the calibration performance of six different methods: Base, ZCoT, DEF, and three reformulated query methods—CG (Counterargument), EX (Explanation), and GO (Goal)—on the PROPAGANDA dataset.}
\label{fig:calibration_propaganda}
\end{figure*}

\begin{figure*}[!ht]
\centering

\includegraphics[width=0.9\linewidth]{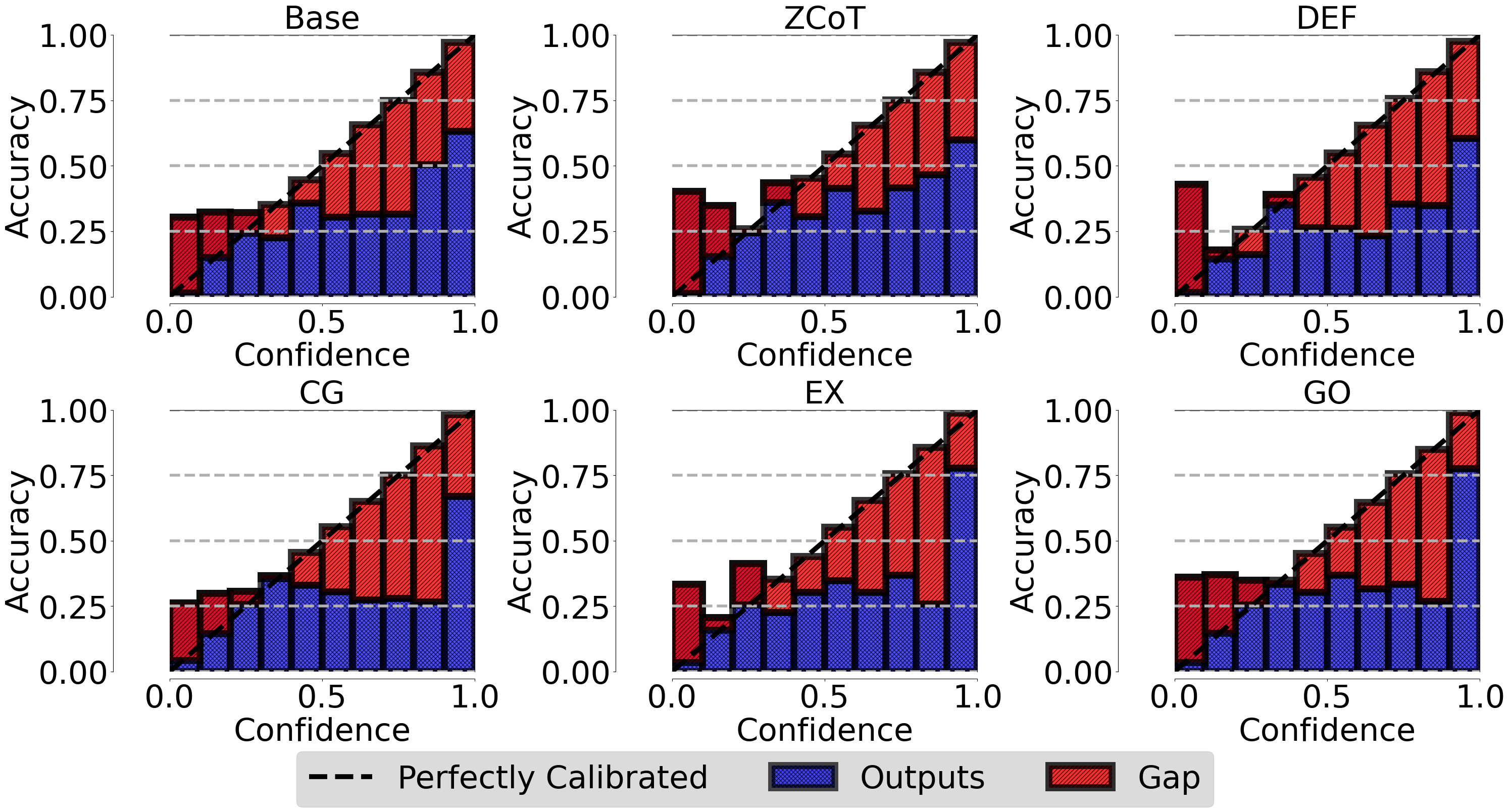}
\caption{Reliability diagrams comparing the calibration performance of six different methods: Base, ZCoT, DEF, and three reformulated query methods—CG (Counterargument), EX (Explanation), and GO (Goal)—on the ARGOTARIO dataset.}
\label{fig:calibration_argotario}
\end{figure*}

\begin{figure*}[!ht]
\centering

\includegraphics[width=0.9\linewidth]{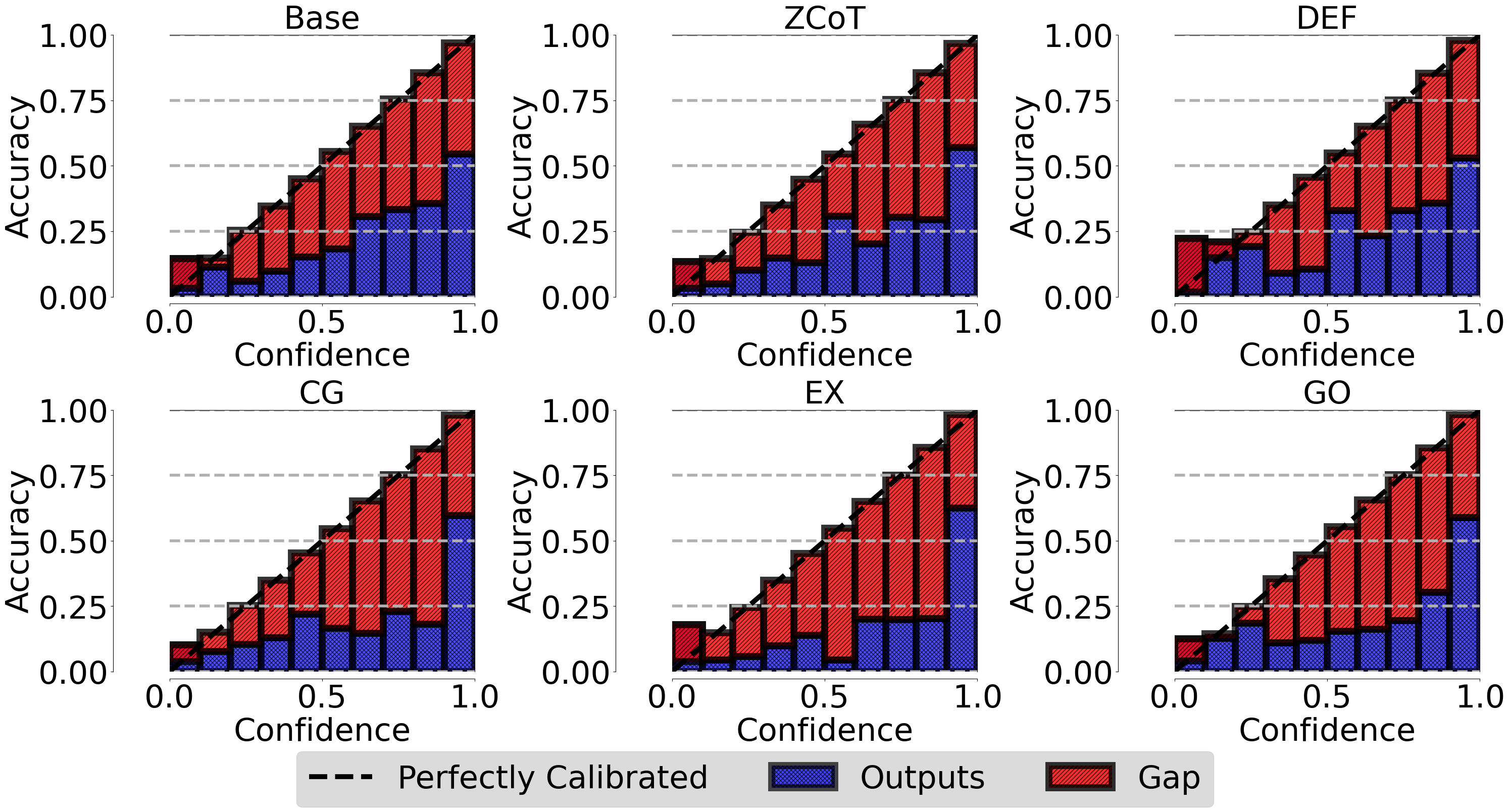}
\caption{Reliability diagrams comparing the calibration performance of six different methods: Base, ZCoT, DEF, and three reformulated query methods—CG (Counterargument), EX (Explanation), and GO (Goal)—on the LOGIC dataset.}
\label{fig:calibration_logic}
\end{figure*}

\begin{figure*}[!ht]
\centering

\includegraphics[width=0.9\linewidth]{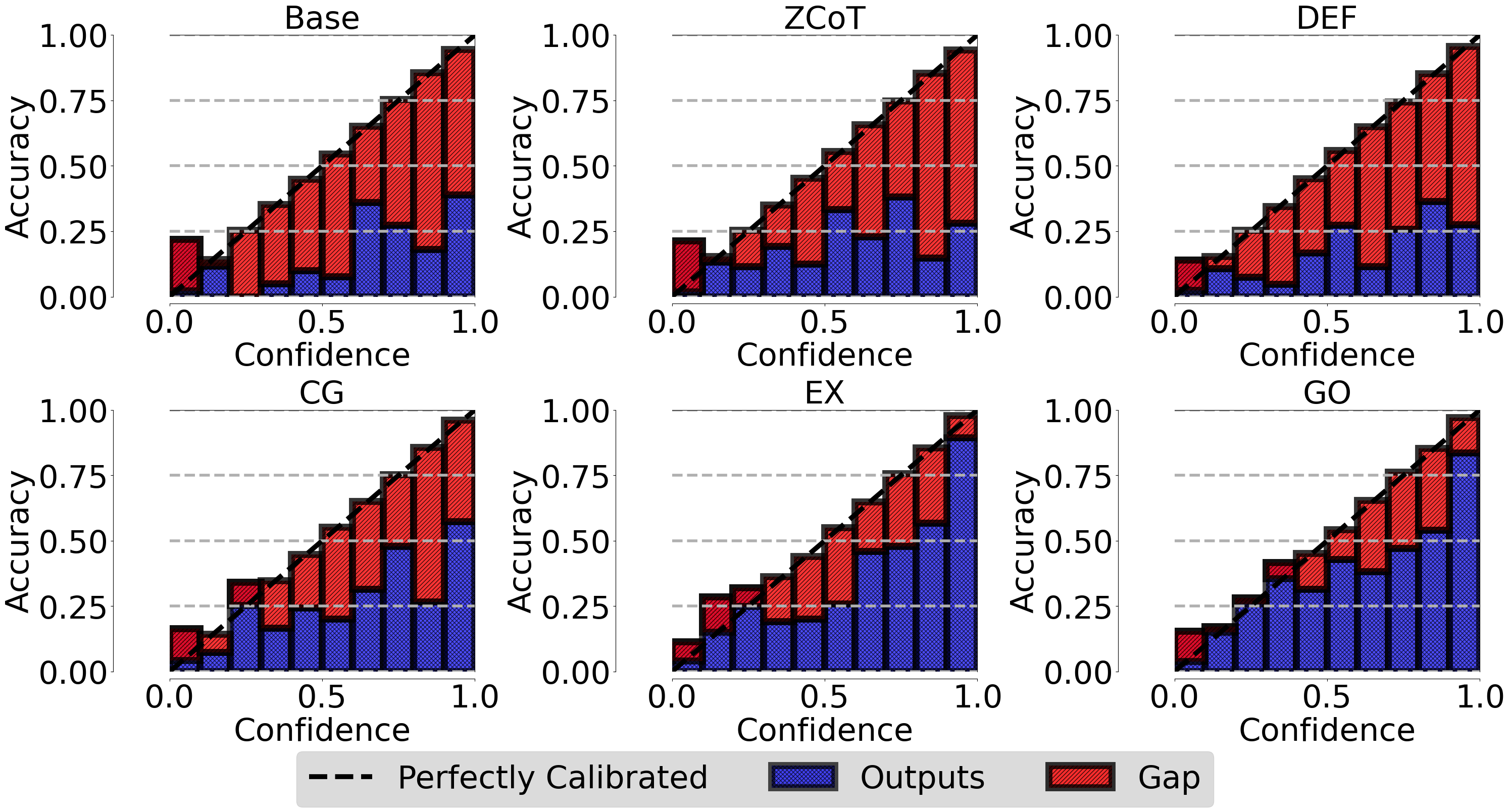}
\caption{Reliability diagrams comparing the calibration performance of six different methods: Base, ZCoT, DEF, and three reformulated query methods—CG (Counterargument), EX (Explanation), and GO (Goal)—on the COVID-19 dataset.}
\label{fig:calibration_covid}
\end{figure*}

\begin{figure*}[!ht]
\centering

\includegraphics[width=0.9\linewidth]{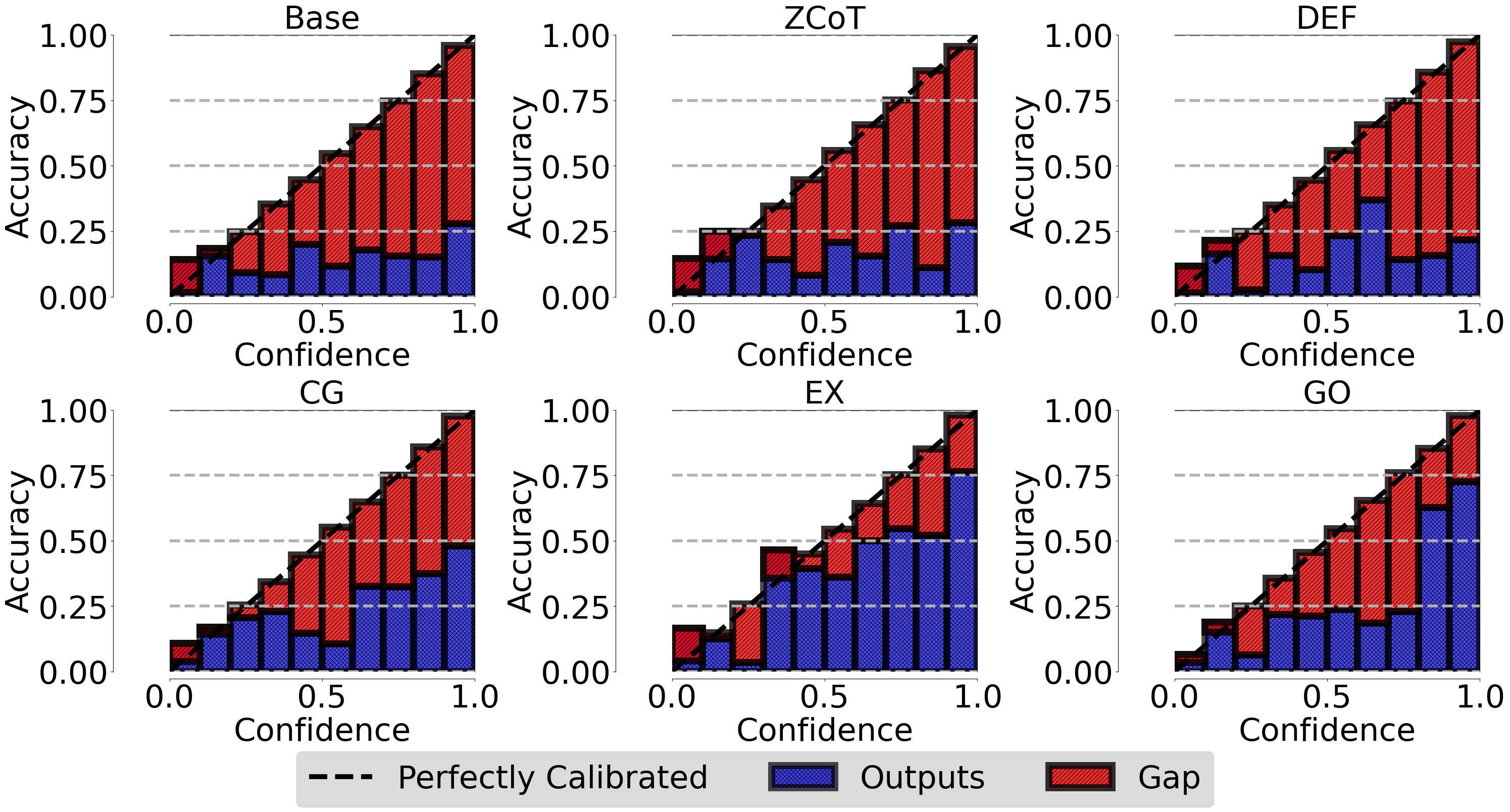}
\caption{Reliability diagrams comparing the calibration performance of six different methods: Base, ZCoT, DEF, and three reformulated query methods—CG (Counterargument), EX (Explanation), and GO (Goal)—on the CLIMATE dataset.}
\label{fig:calibration_climate}
\end{figure*}

The additional results are shown in Figures~\ref{fig:calibration_propaganda} to \ref{fig:calibration_climate}, with an overall comparison between Base and PR across all datasets in Figure~\ref{fig:calibration_all}, highlighting the performance of Base, ZCoT~\cite{kojima2022large}, DEF~\cite{hong2024closer}, and individual queries.

\begin{table*}[!ht]
\centering
\resizebox{\textwidth}{!}{
\begin{tabular}{lccccccccccc}
\toprule
\textbf{Method} & \multicolumn{2}{c}{\textbf{PROPAGANDA}} & \multicolumn{2}{c}{\textbf{ARGOTARIO}} & \multicolumn{2}{c}{\textbf{LOGIC}} & \multicolumn{2}{c}{\textbf{COVID-19}} & \multicolumn{2}{c}{\textbf{CLIMATE}} \\ 
\cmidrule(lr){2-3} \cmidrule(lr){4-5} \cmidrule(lr){6-7} \cmidrule(lr){8-9} \cmidrule(lr){10-11}
 & \textbf{ACC} & \textbf{F1} & \textbf{ACC} & \textbf{F1} & \textbf{ACC} & \textbf{F1} & \textbf{ACC} & \textbf{F1} & \textbf{ACC} & \textbf{F1} \\ \midrule
\textit{Base} & 0.12 & 0.07 & 0.50 & 0.41 & 0.39 & 0.25 & 0.29 & 0.14 & 0.18 & 0.13 \\
\rowcolor{gray!20} \multicolumn{11}{c}{Contextual Augmentation~\cite{sourati2023case}} \\
\textit{+ Counterargument} & 0.13 & 0.08 & 0.48 & 0.40 & 0.35 & 0.20 & 0.21 & 0.13 & 0.12 & 0.07 \\
\textit{+ Explanation} & 0.23 & 0.11 & 0.44 & 0.42 & 0.37 & 0.22 & 0.29 & 0.11 & 0.22 & 0.15 \\
\textit{+ Goal} & 0.19 & 0.11 & 0.49 & 0.41 & 0.35 & 0.21 & 0.29 & 0.15 & 0.18 & 0.12 \\
\rowcolor{gray!20} \multicolumn{11}{c}{Reformulated Query based on \citet{sourati2023case}} \\
\textit{+ Counterargument} & 0.12 & 0.07 & 0.45 & 0.37 & 0.35 & 0.21 & 0.29 & 0.15 & 0.20 & 0.13 \\
\textit{+ Explanation} & 0.21 & 0.09 & 0.47 & 0.37 & 0.35 & 0.21 & 0.29 & 0.16 & 0.25 & 0.17 \\
\textit{+ Goal} & 0.16 & 0.08 & 0.47 & 0.39 & 0.34 & 0.21 & 0.43 & 0.26 & 0.25 & 0.20 \\
\rowcolor{gray!20} \multicolumn{11}{c}{Contextual Augmentation (Ours)} \\
\textit{+ Counterargument} & 0.19 & 0.11 & 0.53 & 0.52 & 0.39 & 0.22 & 0.21 & 0.09 & 0.23 & 0.15 \\
\textit{+ Explanation} & 0.20 & 0.10 & 0.51 & 0.48 & 0.41 & 0.24 & 0.29 & 0.22 & 0.25 & 0.12 \\
\textit{+ Goal} & 0.28 & 0.14 & 0.50 & 0.48 & 0.36 & 0.20 & 0.36 & 0.22 & 0.22 & 0.12 \\
\rowcolor{gray!20} \multicolumn{11}{c}{Reformulated Query from (Ours)} \\
\textit{+ Counterargument} & 0.24 & 0.12 & 0.50 & 0.41 & 0.40 & 0.25 & 0.43 & 0.29 & 0.63 & 0.54 \\
\textit{+ Explanation} & \textbf{0.33} & 0.15 & 0.70 & 0.67 & \textbf{0.46} & \textbf{0.32} & 0.57 & 0.44 & \textbf{0.70} & \textbf{0.65} \\
\textit{+ Goal} & \textbf{0.33} & \textbf{0.16} & \textbf{0.74} & \textbf{0.72} & 0.44 & \textbf{0.32} & \textbf{0.64} & \textbf{0.47} & 0.58 & 0.49 \\
\bottomrule
\end{tabular}
}
\caption{Performance comparison based on different contextual augmentation prompt methods and their resulting reformulated queries across five datasets using the \texttt{gpt-3.5-turbo} model. The prompt used to generate reformulated queries is the same for all cases. F1: Macro-F1 score.}
\label{tab:comparison_all}
\end{table*}

\renewcommand{\arraystretch}{1.2} 
\begin{table*}[ht]
\centering
\resizebox{\textwidth}{!}{
\begin{tabular}{@{}p{0.3\textwidth} p{0.7\textwidth} @{}}
\toprule
\textbf{Method} & \textbf{Prompt} \\
\midrule
\textit{ZCoT \cite{kojima2022large}} &
Your task is to detect a fallacy in the Text. The label can be `Appeal to Emotion', `Faulty Generalization', `Red Herring', `Ad Hominem', and `Irrelevant Authority'.\par
Please detect a fallacy in the Text. \textbf{Let's think step by step}. \\

\midrule
\textit{DEF \cite{hong2024closer}} &
Your task is to detect a fallacy in the Text. The label can be `Appeal to Emotion', `Faulty Generalization', `Red Herring', `Ad Hominem', and `Irrelevant Authority'.\par
1. Appeal to Emotion: This fallacy tries to arouse non-rational sentiments within the intended audience in order to persuade.\par
2. Faulty Generalization: The argument uses a sample which is too small, or follows falsely from a sub-part to a composite or the other way round.\par
3. Red Herring: This argument distracts attention to irrelevant issues away from the thesis which is supposed to be discussed.\par
4. Ad Hominem: The opponent attacks a person instead of arguing against the claims that the person has put forward.\par
5. Irrelevant Authority: While the use of authorities in argumentative discourse is not fallacious inherently, appealing to authority can be fallacious if the authority is irrelevant to the discussed subject.\par
Please detect a fallacy in the Text. \\
\bottomrule
\end{tabular}
}
\caption{Comparison of prompt methods for fallacy classification using ZCoT~\cite{kojima2022large} and DEF~\cite{hong2024closer} approaches. For ZCoT, the prompt "Let's think step by step" is added. DEF uses definitions for each class, which are included in the prompt. The fallacy classes in the DEF prompt are from the ARGOTARIO dataset. If a query is included, the last sentence should be modified to "Please detect a fallacy in the text based on the query."}
\label{tab:prompt-baseline}
\end{table*}

\begin{table*}[!ht]
\centering
\renewcommand{\arraystretch}{1.2} 
\begin{tabular}{@{}p{0.3\textwidth} p{0.20\textwidth} m{0.40\textwidth} @{}}
\toprule
\textbf{Method} & \textbf{Perspective} & \textbf{Prompt} \\
\midrule
\textit{CA \cite{sourati2023case}} & CG & \raggedright Represent the counterargument to the text. \arraybackslash \\
\cmidrule{2-3}
& EX & \raggedright Analyze the text. \arraybackslash \\
\cmidrule{2-3}
& GO & \raggedright Express the goal of the text. \arraybackslash \\
\midrule
\textit{CA (Ours)} & CG & \raggedright I'll give you some texts. The texts can be question and answer pairs or sentences. The text contains one of following logical fallacies: \{fallacy\_class\} $\,$. Represent the counterargument to the text.
 \arraybackslash \\
\cmidrule{2-3}
& EX & \raggedright I'll give you some texts. The texts can be question and answer pairs or sentences. The text contains one of following logical fallacies: \{fallacy\_class\}. Analyze the text. \arraybackslash \\
\cmidrule{2-3}
& GO & \raggedright I'll give you some texts. The texts can be question and answer pairs or sentences. The text contains one of following logical fallacies: \{fallacy\_class\}. Express the goal of the text. \arraybackslash \\
\midrule
\textit{Reformulated Query} & CG & \raggedright I'll give you some texts and text's counterarguments. The texts can be question and answer pairs or sentences. Create one question for each text that analyzes the text based on counterarguments rather than directly asking what logical fallacy is. \arraybackslash \\
\cmidrule{2-3}
& EX & \raggedright I'll give you some texts and text's explanations. The texts can be question and answer pairs or sentences. Create one question for each text that analyzes the text based on explanations rather than directly asking what a logical fallacy is. \arraybackslash \\
\cmidrule{2-3}
& GO & \raggedright I'll give you some texts and text's goals. The texts can be question and answer pairs or sentences. Create one question for each text that analyzes the text based on goals rather than directly asking what a logical fallacy is. \arraybackslash \\
\bottomrule
\end{tabular}
\caption{Comparison of prompt methods for contextual augmentations (CA) and reformulated queries, illustrating the approaches by \citet{sourati2023case} and our method. Both CA and reformulated queries are generated using the \texttt{gpt-3.5-turbo-instruct} model. CG: Counterargument, EX: Explanation, and GO: Goal. \{fallacy\_class\} refers to the classes in the dataset to which the sample (sentence) belongs.}
\label{tab:prompt}
\end{table*}


\end{document}